\documentclass[10pt,twocolumn,letterpaper]{article}

\usepackage[pagenumbers]{cvpr} % To force page numbers, e.g. for an arXiv version

\usepackage[dvipsnames,table]{xcolor}
\usepackage{supertabular}

\usepackage[T1]{fontenc}
\usepackage{multirow}
\usepackage{bm}

\definecolor{cvprblue}{rgb}{0.21,0.49,0.74}
\usepackage[pagebackref,breaklinks,colorlinks,citecolor=cvprblue]{hyperref}

\title{ Dynamic Loss Decay based Robust Oriented Object Detection \\ on Remote Sensing Images with Noisy Labels }

\def\author{
Guozhang Liu,  Ting Liu,  Mengke Yuan,  Tao Pang \\ Guangxing Yang,  Hao Fu,  Tao Wang\thanks{Corresponding author, wangtao@piesat.cn},  Tongkui Liao\\ 
\\        
Piesat Information Technology\\
{\tt\small liuguozhang@aliyun.com} \\
{\tt\small \{tliu068 mannix.yuan pangtao0925 guaxingyang\}@gmail.com} \\
{\tt\small \{fuhao\_zn wangtao liaotongkui\}@piesat.cn} \\
}
\begin{document}
\maketitle
\begin{abstract}

The ambiguous appearance, tiny scale, and fine-grained classes of objects in remote sensing imagery inevitably lead to the noisy annotations in category labels of detection dataset. However, the effects and treatments of the label noises are underexplored in modern oriented remote sensing object detectors. To address this issue,  we propose a robust oriented remote sensing object detection method through dynamic loss decay (DLD) mechanism, inspired by the two phase ``early-learning'' and ``memorization'' learning dynamics of deep neural networks on clean and noisy samples. To be specific, we first observe the end point of early learning phase termed as $\textbf{EL}$, after which the models begin to memorize the false labels that significantly degrade the detection accuracy. Secondly, under the guidance of the training indicator, the losses of each sample are ranked in descending order, and we adaptively decay the losses of the top K largest ones (bad samples) in the following epochs. Because these large losses are of high confidence to be calculated with wrong labels. Experimental results show that the method achieves excellent noise resistance performance tested on multiple public datasets such as HRSC2016 and DOTA-v1.0/v2.0 with synthetic category label noise.
Our solution also has won the 2nd place in the "fine-grained object detection based on sub-meter remote sensing imagery" track with noisy labels of 2023 National Big Data and Computing Intelligence Challenge.
\end{abstract}

\begin{figure}[ht]
\centering
\includegraphics[width=1\linewidth]{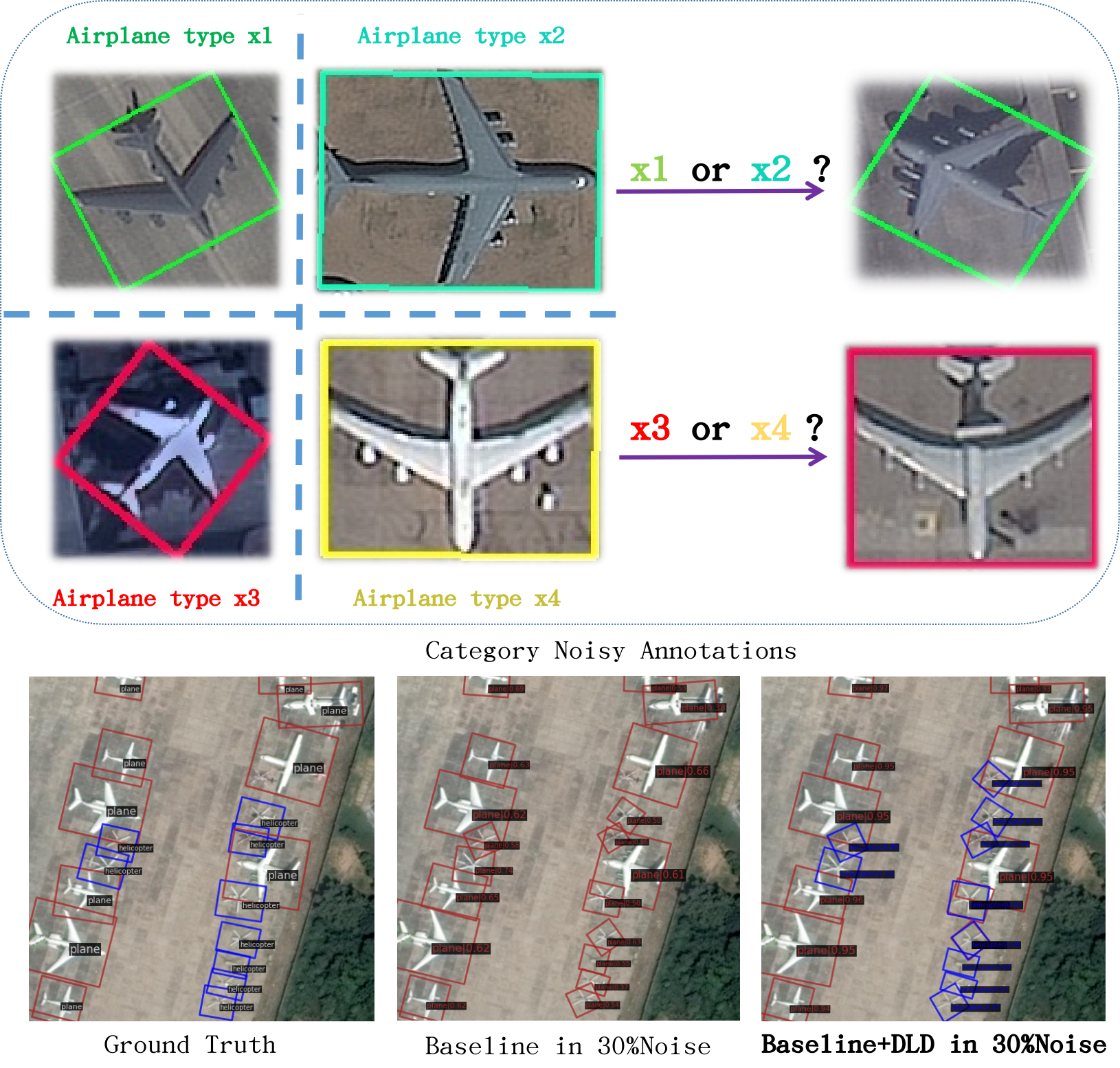}
\caption{The first row illustrates the difficulty of annotating the fine-grained types of planes in remote sensing images with similar appearances. The second row shows one example image with correct reference ground truth annotations in DOTA-v1.0 dataset on the left. The red and blue oriented bounding boxes indicate ``plane'' and ``helicopter''. We train baseline ORSOD model and baseline + \textbf{DLD(ours)} with synthesized 30\% noisy category labels. Their detection results are visualized in the middle and right respectively, baseline result contains several false classification instances. Baseline with \textbf{DLD} generates more accurate results.}
\label{fig:blur_images}
\end{figure}

% \begin{figure}[ht]
% \centering
% \includegraphics[width=0.9\linewidth]{graphs/contrast.png}
% \caption{Visulization of of ground truth from DOTAv1.0, and results from Lsknet-Tiny trained under 30\% category noisy labels. The last column shows the results from our methon(DLD).}
% \label{fig:infer_images}
% \end{figure}

\section{Introduction}
\label{sec:intro}
\quad Extensive researches have been devoted to recognize objects in remote sensing imagery and locate them with more precise oriented bounding boxes, i.e. oriented remote sensing object detection (ORSOD), which is of great interest in both computer vision and remote sensing community. Most of them focus on improving performance with cutting-edge detection frameworks, such as anchor-based one/two-stage detection\cite{yang2021r3det,han2021align,ding2019learning,xie2021oriented}, anchor-free point-based detection~\cite{guo2021beyond,li2022oriented} and DETR-based detection~\cite{dai2022ao2,zeng2023ars-detr}. Some other works pay attention to design network components, like augmented backbones~\cite{rvsa,Li_2023_ICCV}, elaborated loss functions and angle coders~\cite{yang2021rethinking,yang2021learning,yang2021dense,yu2023phase}, effective label assigners~\cite{huang2022general,xu2023dynamic}. However, few ORSOD methods take notice of the ubiquitous and detrimental noisy annotations in current dataset, since the data set with high 
quality, low cost, and large scale can not be simultaneously achieved. 

The noisy annotations have raised concerns in training image classification, segmentation and object detection models for decades. A series of robust loss functions~\cite{englesson2021generalized,liu2020early,ahn2023sample} and clean sample selection methods~\cite{jiang2018mentornet,han2018co,feng2023ot} are proposed to alleviate the effects of noisy labels in image classification. Like-wisely, in more challenging image segmentation task, annotation noises resistant methods~\cite{jacob2021disentangling,wang2020noise,liu2022adaptive,li2023semi} are designed to avoid the degradation. More related object detection methods~\cite{zhang2019learning,li2020towards,xu2021training, bernhard2021correcting,hu2022probability,liu2022robust,wu2023spatial} pay more attention to inaccurate horizontal bounding boxes or mixture of categorical and positional annotation noises. None of them addresses the category annotation noises in training ORSOD model. In practice, distinguishing the specialized type of small size, dense arranged objects (such as ships and planes) with remote sensing imagery of sub-meter spatial resolution is difficult even for the experts as shown in the first row of Figure~\ref{fig:blur_images}. The inter-class discrepancy is small for fine-grained plane types. It isn't surprising that the datasets contains many category label noises. In contrast to the inaccurate bounding box, the absolute wrong class label deserves particularly consideration to acquire more robust ORSOD model.

To address the challenge in training model with category noisy labels,  we propose a robust oriented remote sensing object detection method through dynamic loss decay  mechanism, inspired by the two phase ``early-learning'' and ``memorization'' learning dynamics of deep neural networks (DNNs) on clean and noisy samples~\cite{zhang2021understanding,arpit2017closer}. Although, \cite{zhang2021understanding} points that DNNs can easily fit a random labeling of the training data, the quantitative differences are demonstrated in DNNs optimization on clean and noisy data~\cite{arpit2017closer}. The DNNs learn simple consistently shared patterns among training samples before memorizing irregular false labels. The learning dynamics of DNNs is adopted in training classification~\cite{liu2020early} and segmentation models~\cite{liu2022adaptive} with noisy annotations. Imposing early-learning regularization~\cite{liu2020early} or correcting probable false labels ~\cite{liu2022adaptive} in early learning stage are proved to be effective. 

Similarly, in training ORSOD model with noisy category labels, we have observed consistent accelerated and decelerated improvement dynamics of both \textbf{mAP} (measured using model output and ground truth), and top-1 accuracy \textbf{ACC} (measured using model output and noisy annotations), which can be used to find the endpoint of early-learning phase represented by \textbf{EL}. Specifically, we first identify \textbf{EL} through monitoring the second-derivative of \textbf{ACC} curve during training. The accuracy curves of models trained on category labels with different proportion of noises share the same trend and \textbf{EL}. Secondly, the dynamic loss decay begins in the memorization phase.  Since the larger loss value means the higher probability of being calculated with false labels, the overall loss computation is divided into two parts, the top K largest samples losses and the rest samples losses. We adaptively decrease the weight of the largest top K samples losses which contains the most false category labels in the following epochs. Experimental results corroborate that our method is robust to categorical annotation noises in ORSOD, and effectively decreases the model performance degradation when training on manually contaminated public dataset HRSC2016 and DOTA-v1.0/v2.0. Our solution also has won the 2st place in the ``fine-grained object detection based on sub-meter remote sensing imagery'' track with artificial class noise annotations of 2023 National Big Data and Computing Intelligence Challenge (NBDCIC 2023)~\cite{nbdcic2023}. In summary, our contributions are:
\begin{itemize}
    \item We propose the first robust ORSOD method against categorical annotation noises through dynamic loss decay mechanism.
    \item We identify the effective early-learning phase endpoint \textbf{EL} in training accuracy curves through theoretical and experimental analysis for ORSOD. 
    \item We validate the superiority of proposed method in both common ORSOD benchmarks and competitive NBDCIC 2023.
\end{itemize}

\begin{figure*}[h]
    \centering
    \begin{subfigure}[b]{0.24\textwidth}
	\begin{minipage}[t]{\linewidth}
		\includegraphics[width=1.6in]{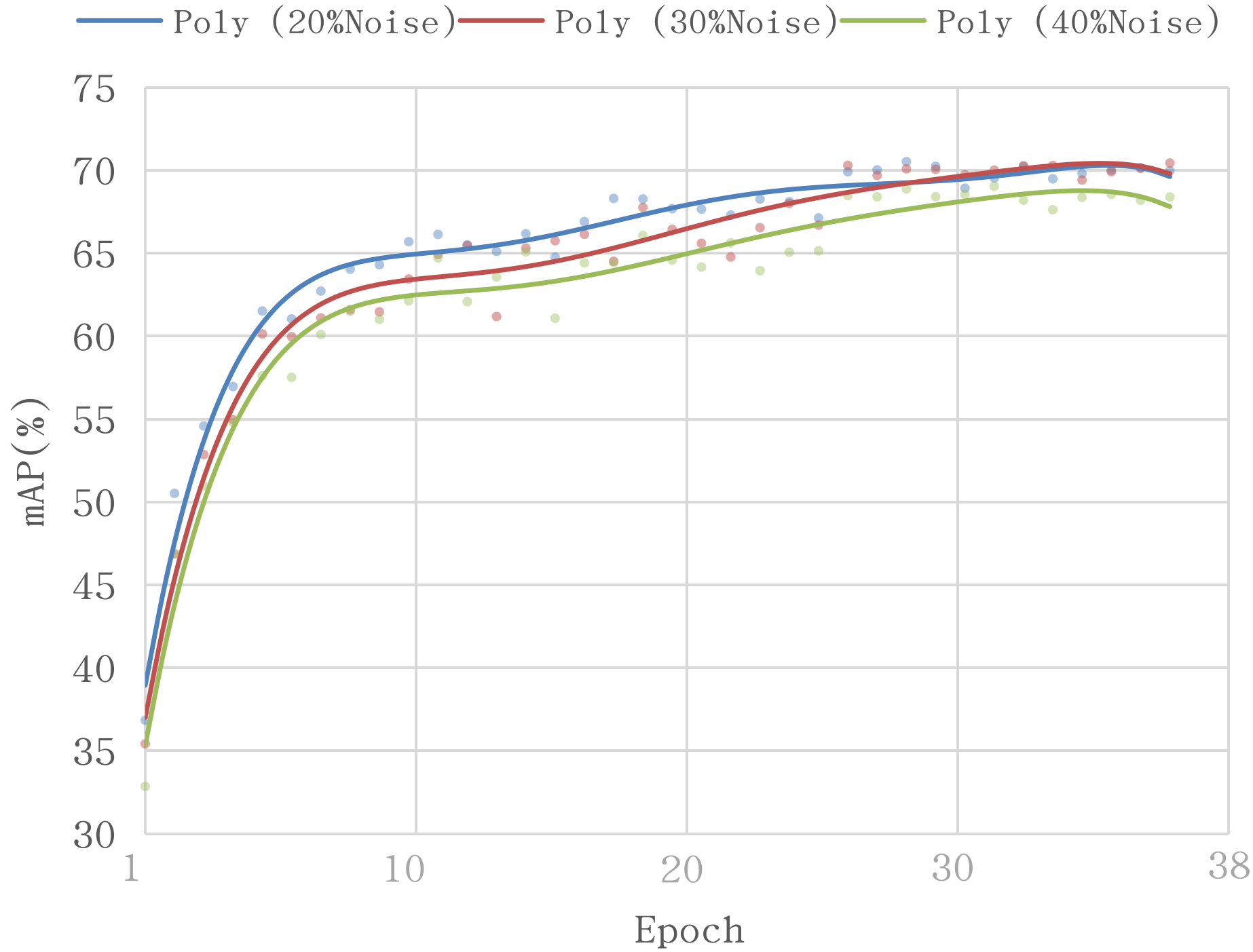} 
            \label{val_map}
            \caption{ \textbf{mAP} }
	\end{minipage}
    \end{subfigure}
    \begin{subfigure}[b]{0.24\textwidth}
	\begin{minipage}[t]{\linewidth}
		\includegraphics[width=1.6in]{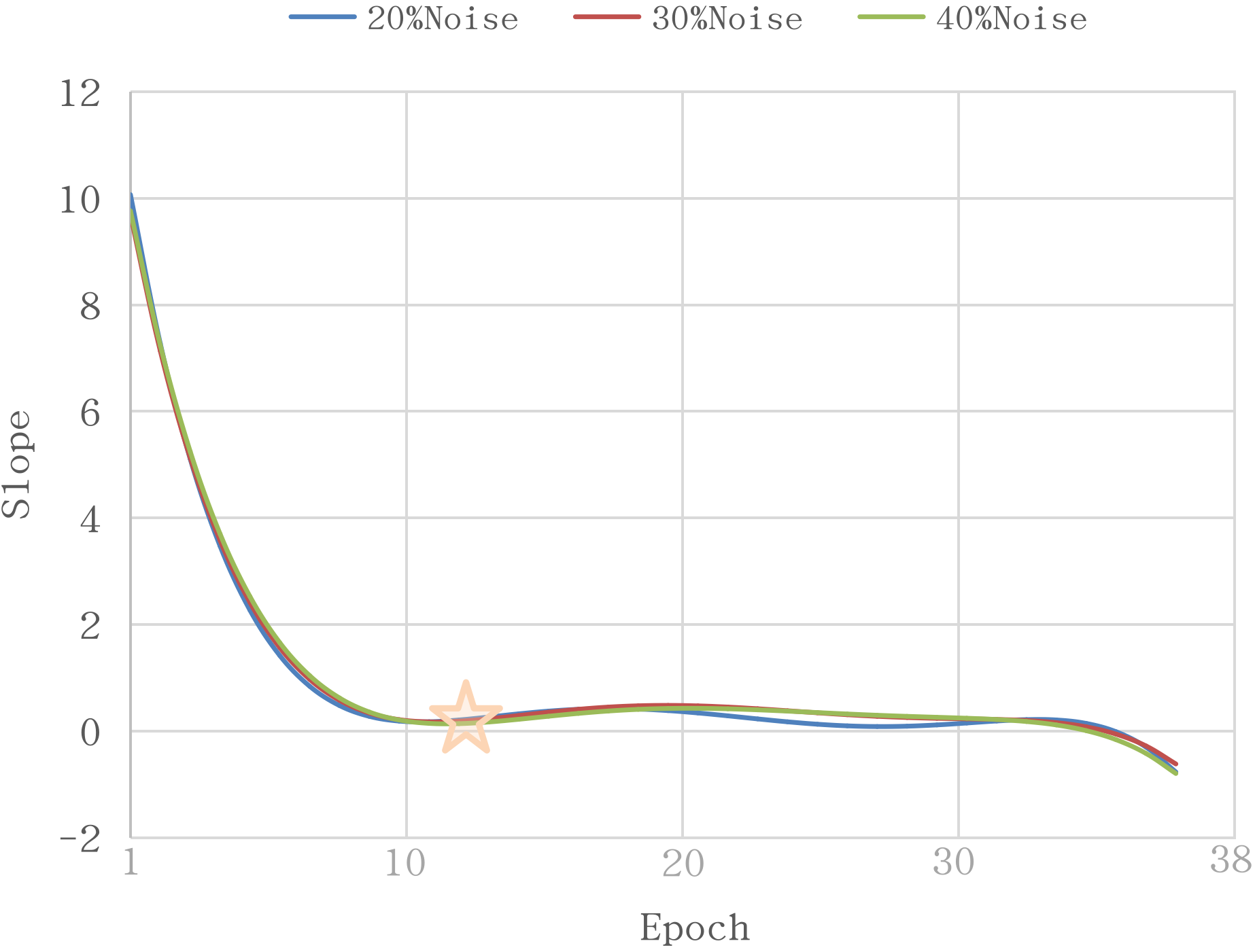}
            \label{val_map_slope}
            \caption{Slope of \textbf{mAP}}
	\end{minipage}
    \end{subfigure}
    \begin{subfigure}[b]{0.24\textwidth}
	\begin{minipage}[t]{\linewidth}
		\centering
		\includegraphics[width=1.6in]{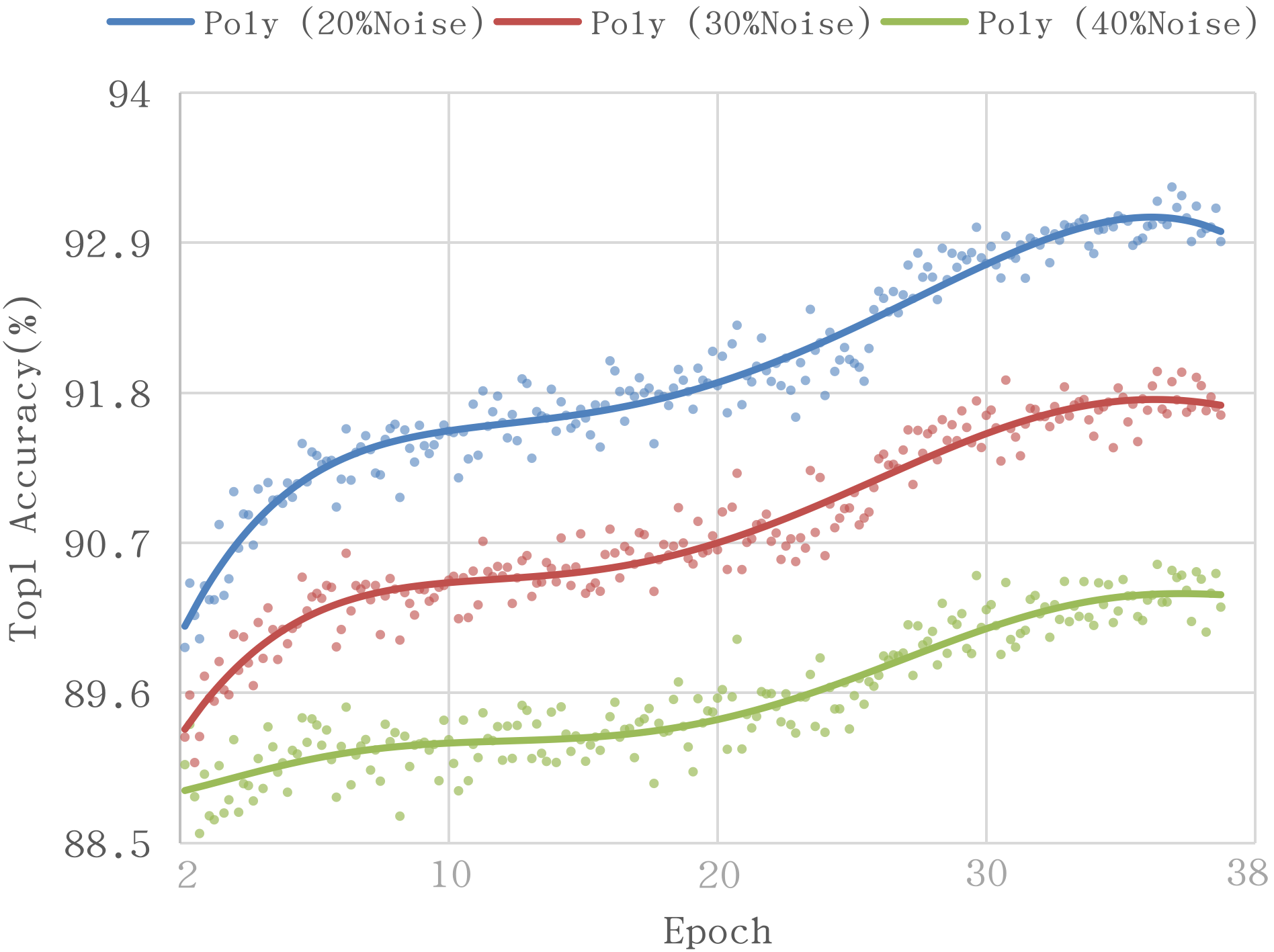}
            \label{fitted_acc}
            \caption{\textbf{ACC}}
	\end{minipage}
    \end{subfigure}
    \begin{subfigure}[b]{0.24\textwidth}
	\begin{minipage}[t]{\linewidth}
		\centering
		\includegraphics[width=1.6in]{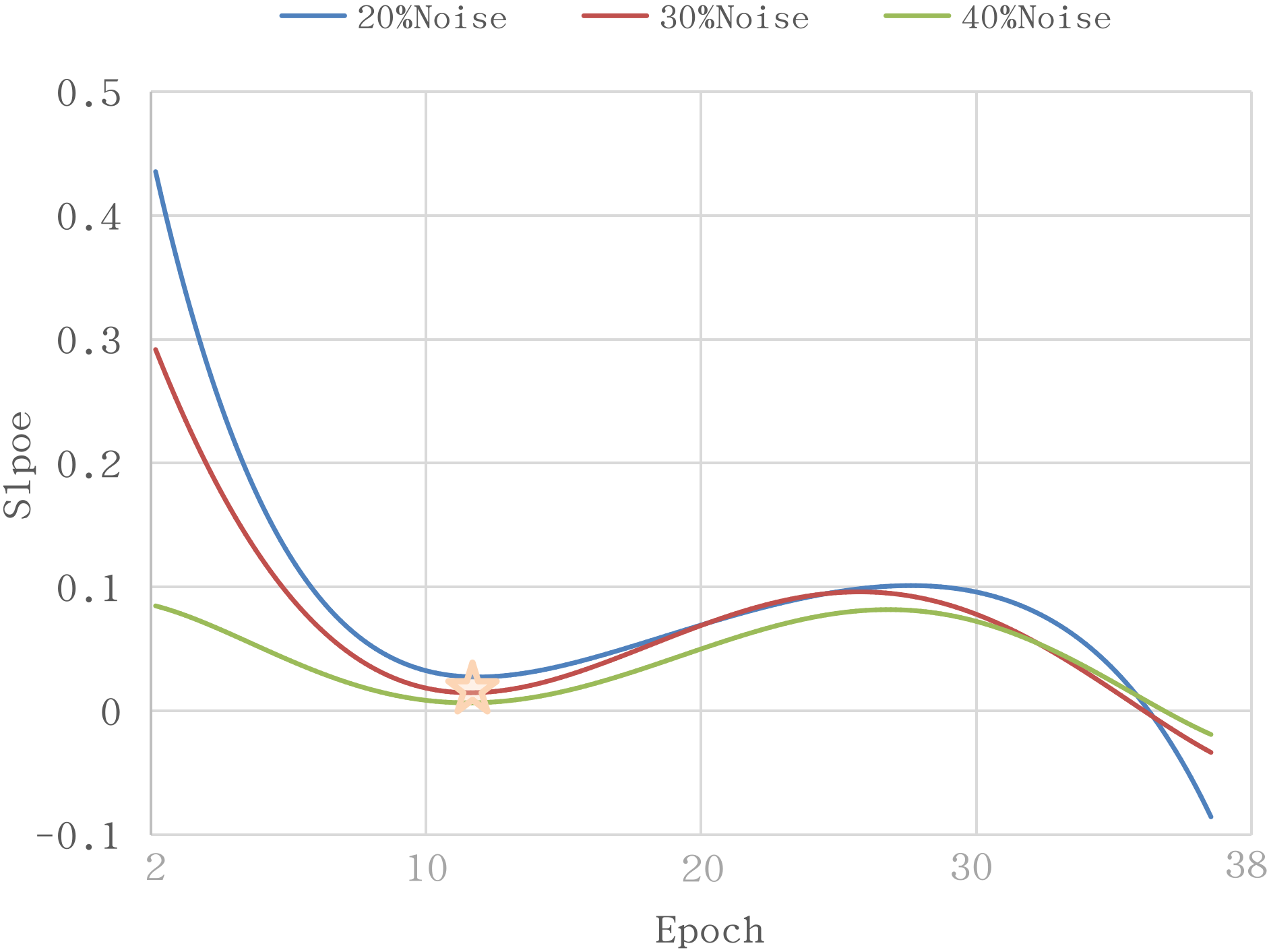}
            \label{acc_slope}
            \caption{Slope of \textbf{ACC}}
	\end{minipage}
    \end{subfigure}
    \caption{The dynamics of two measurements mean average precision (\textbf{mAP}) and top 1 accuracy (\textbf{ACC}), acquired by the  Oriented R-CNN with LSKNet-Tiny backbone. The experiments are conducted on DOTA-v1.0 dataset contaminated with different level of category noises (20\%, 30\%, and 40\%). The \textbf{mAP} is calculated between model output and clean GT category labels. The \textbf{ACC} of the model output is referenced with noisy category labels. The star in (b)(d) represents the early-learning endpoint.}
    \label{fig:fitted_map_acc_curve}
\end{figure*}

\section{Related Works}
\label{sec:related_works}

\subsection{Oriented Remote Sensing Object Detection}
\quad Most object detection methods commonly utilize horizontal bounding box (HBB) to localize general objects. Considering the severe overlapping by using HBB to represent bird-view remote sensing objects, oriented bounding box (OBB) representation is more accurate with extra direction. There are both similarities and discrepancies between HBB based general object detection and  OBB based remote sensing object detection. On the one hand, inspired by HBB based object detection framework, representative anchor-based one-stage (R$^3$Det~\cite{yang2021r3det}, S$^2$A-Net~\cite{han2021align}) and two-stage (ROI-Transformer~\cite{ding2019learning}, Oriented-RCNN~\cite{xie2021oriented}), as well as anchor-free point-based (CFA~\cite{guo2021beyond}, Oriented RepPoints~\cite{li2022oriented}) and DETR-based (A$^2$O-DETR~\cite{dai2022ao2}, ARS-DETR~\cite{zeng2023ars-detr}) OBB remote sensing object detectors are proposed. On the other hand, to accommodate the additional rotation variance, backbones are augmented with rotation varied-size window attention (RVSA~\cite{rvsa}) and large selective kernel (LSKNet~\cite{Li_2023_ICCV}), label assigners (GGHL~\cite{huang2022general}, DCFL~\cite{xu2023dynamic}) are integrated to fulfill effective label assignment, as well as loss functions (GWD~\cite{yang2021rethinking}, KLD~\cite{yang2021learning}) and angle coders (DCL~\cite{yang2021dense}, PSC~\cite{yu2023phase}) have been designed to address the angular boundary discontinuity and square-like problem. Interested readers can refer to more comprehensive literature in the survey~\cite{wang2023oriented}. 

\subsection{Learning with Noisy Labels}
\quad Note-worthily, the acquisition and annotation cost of remote sensing data is much higher than ground-based data. Training models with incomplete, inexact, and inaccurate supervision which collectively referred as weakly supervised  in ~\cite{zhou2018brief} is practical and desirable. Researchers are highly motivated to construct noise-resistant models in classification, segmentation, and object detection tasks.

\textbf{Classification}. Various tactics such as robust loss functions (GJS~\cite{englesson2021generalized}, SLC~\cite{ahn2023sample}, ELR~\cite{liu2020early}), sample selection(MentorNet~\cite{jiang2018mentornet}, co-teaching~\cite{han2018co}, OT-filter~\cite{feng2023ot}) are designed to handle categories label noises in classification. In remote sensing, \cite{jiang2020multilayer,tu2020hyperspectral,zhang2023triple}
address the issue of noisy labels in hyperspectral image classification. Specifically, both ELR~\cite{liu2020early} and our proposed object category noise treatment are built upon the learning dynamics of deep neural networks (DNNs) on clean and noisy samples as disclosed in \cite{zhang2021understanding} and \cite{arpit2017closer}. The DNNs have been observed to first fit clean labels during ``early learning phase", and then memorize false labels in ``memorization phase". 

\textbf{Segmentation}. Image segmentation performs dense pixel-wise classification, and is more challenging. Both in the medical and remote sensing domain, the noisy annotations are ubiquitous on account of weary labeling and expertise requirement. Medical segmentation methods improve the robustness by modeling human annotation errors~\cite{jacob2021disentangling}, adopting regularization term~\cite{wang2020noise}, etc. Note that in contrast to noisy label classification, not all semantic categories share the synchronous learning dynamics in segmentation, ADELE~\cite{liu2022adaptive} separately correct noisy label for each category. In \cite{li2023semi}, the authors propose a semi-supervised segmentation method to handle both incomplete and inaccurate labels via multiple diverse learning groups. 

\textbf{Object detection}.
The complexity of object detection with noisy annotations lies in simultaneously dealing with possible categorical and positional noises. Much effort has been devoted to robust HBB based object detection~\cite{zhang2019learning,li2020towards,xu2021training, bernhard2021correcting,hu2022probability,liu2022robust,wu2023spatial}, while we can hardly found studies on OBB based noise-resistant remote sensing object detection. Object detection with noisy labels is closely related to weakly-supervised object detection (WSOD)~\cite{bilen2016weakly}. The commonly WSOD setting is train detectors with image-level labels and follows multiple instance learning (MIL) pipeline to joint optimization of object appearance and object region in positive bags. SD-LocNet~\cite{zhang2019learning} utilizes adaptive sampling to select reliable object instances to filter the noisy initialized object proposals in WSOD. In \cite{li2020towards}, the mixture of label noise and bounding box noise are handled with alternating noise correction and model training. As well, MRNet~\cite{xu2021training} leverages a meta-learning frame work to dynamically weight incorrect categories labels and refine imprecise bounding boxes in model training. OA-MIL~\cite{liu2022robust}  focuses on inaccurate bounding box, and selects more accurate HBB with instance generation and extension.
Similarly, SSD-Net~\cite{wu2023spatial} only deals with noisy HBB, and refines them in a self-distillation fashion. For HBB based remote sensing object detection, \cite{bernhard2021correcting} proposes a co-correction scheme to align noisy object locations with learned class activation map, and \cite{hu2022probability} designs probability deferential noise filter to identify and revise wrong class label.

\section{Our Approach}
\label{sec:methods}
\quad In this section, we present a straightforward yet innovative robust ORSOD method via dynamic loss decay mechanism, which is capable of handling category-noisy annotations effectively. Grounding on the two-phase learning dynamics of DNNs~\cite{arpit2017closer}, the key of DLD is identifying the end point of the early learning stage, and begin to mitigate the interference of noisy labels by gradually reducing the weight of top ranked sample losses, which are most probably measured with false references. We will detail the preliminary of learning dynamics of DNNs with noisy labels, how to find the end point of early-learning phase in ORSOD, and dynamic loss decay for robust ORSOD.  

\begin{figure*}[ht]
\centering
\includegraphics[width=1\linewidth]{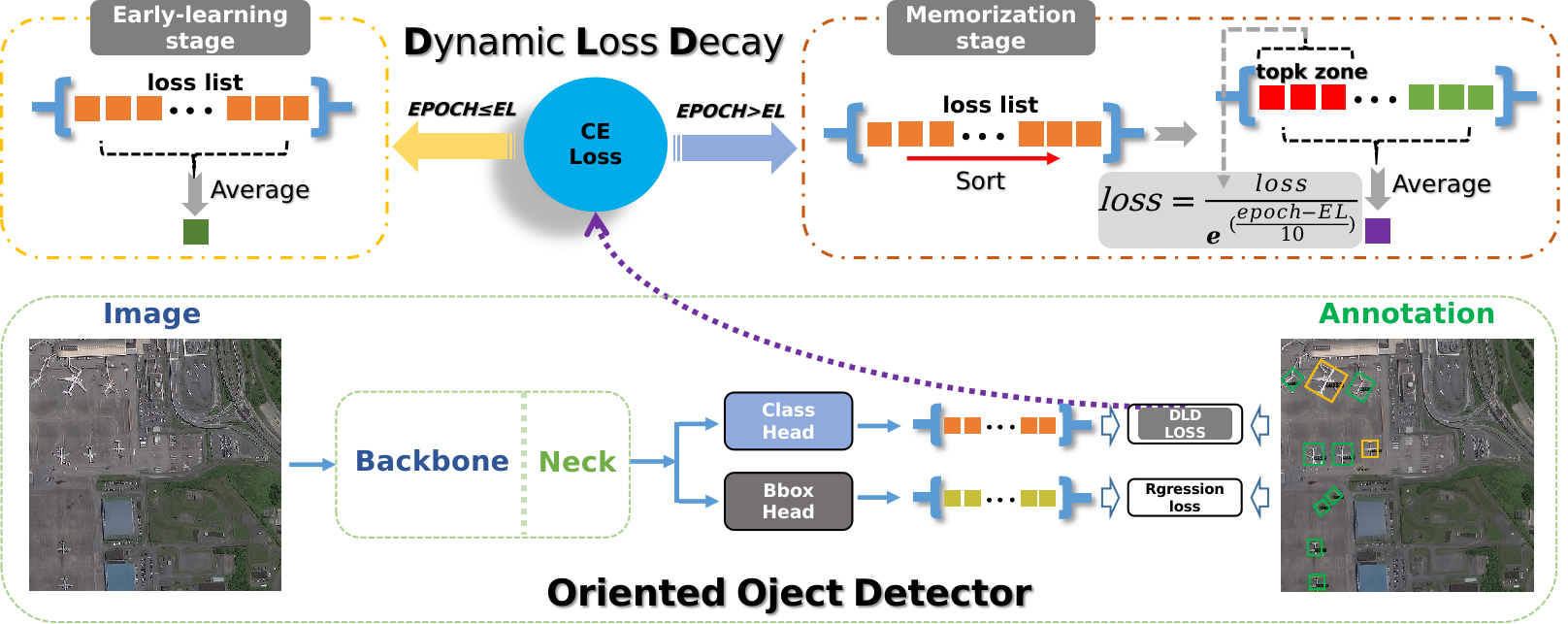}
\caption{The overview training process of \textbf{DLD}. The bottom part illustrates structure of an Oriented Object Detector, the upper part shows the conceptual illustration of DLD based on early-learning stage and memorization stage theory.}
\label{fig:overview}
\end{figure*}

\subsection{Preliminary}
\label{sec:Pre}
\quad In ELR~\cite{liu2020early}, the authors theoretically interpret the learning dynamics of DNNs from gradient analysis. Considering the  classification problem of training DNNs with $N$ samples $\{\bm{x}_i,\bm{a}_i\}_{i=1}^N$ in $C$ classes, where $\bm{x}_i \in \mathcal{R}^d$ is the $i$-th sample and $\bm{a}_i \in \{0,1\}^C$ is the corresponding one-hot annotation vector. The DNNs $\mathcal{D}_{\Theta}$ parameterized with $\Theta$ encodes $\bm{x}_i$ into $C$-dimensional feature $\mathcal{D}_{\Theta}(\bm{x}_i) \in \mathcal{R}^C$, and we can acquire the conditional probability prediction $\bm{p}_i = \mathcal{S}(\mathcal{D}_{\Theta}(\bm{x}_i))$ with softmax function $\mathcal{S}$. The cross-entropy loss \eqref{eq:loss} and gradient with respect to $\Theta$ \eqref{eq:gradient} can be formulated as :
\begin{align} 
\mathcal{L}_\text{CE}(\Theta) &:=-\frac{1}{N}\sum_{i=1}^N  \sum_{c=1}^C a^c_i \log p^c_i \label{eq:loss}\\ 
\nabla \mathcal{L}_\text{CE}(\Theta) &= \frac{1}{N}\sum_{i=1}^N \nabla \mathcal{D} (\Theta | \bm{x}_i ) \left( \bm{p}_i-\bm{a}_i \right)
\label{eq:gradient}
\end{align}
where $\nabla \mathcal{D} (\Theta | \bm{x}_i )$ is the Jacobian matrix of DNNs encoding for $i$-th sample $\bm{x}_i$ with respect to $\Theta$.  Therefore, the contribution of $i$-th sample $x_i$ to the gradient of class $c$ is $\nabla \mathcal{L}^c_\text{CE}(\Theta) = \nabla \mathcal{D} (\Theta | \bm{x}_i ) \left( p^c_i-a^c_i \right)$. If $x_i$ truly belongs to class $c$, i.e. $a^c_i = 1$,  the gradient descent will forward to the right direction $\nabla \mathcal{D} (\Theta | \bm{x}_i )$, otherwise the wrong annotation will push the update of $\Theta$ to the opposite direction.

The learning dynamics of DNNs reflects the characteristics of gradient descent with noisy labels. In early-learning stage, since correct labels are in majority and the gradient descent is well correlated with optimal direction, we can expect accelerated accuracy increment. Once the magnitude of noisy gradient dominates the update, the DNNs will memorize (overfit) the noise label, and the improvement of accuracy will decelerate.

\subsection{End Point of Early-learning in ORSOD}
\label{sssec:early-learning}

\quad The existence of the turning point between early-learning phase and memorization phase, in training classification DNNs with noisy labels, has been theoretically proven in ELR~\cite{liu2020early}. The point has also been validated in the learning dynamics of image segmentation DNNs training with noisy pixel annotations in ADELE~\cite{liu2022adaptive}. Similarly, we experimentally demonstrate that the same end point of early-learning consistently occurs in training ORSOD model with different level of noisy category labels.    

Specifically, we monitor the dynamics of two measurements mean average precision (\textbf{mAP}) and top 1 accuracy (\textbf{ACC}), acquired by the representative ORSOD method Oriented R-CNN with LSKNet-Tiny backbone~\cite{Li_2023_ICCV}. The experiments are conducted on the well-known DOTA-v1.0 dataset contaminated with different level of category noises (20\%, 30\%, and 40\%), and experimental details is given in Section \ref{sec:experiments}. The \textbf{mAP} is calculated between model output and clean ground truth (GT) category labels. In practice, we have no access to clean GT in real-world application, the model output \textbf{ACC} with reference to noisy category labels can serve as a surrogate for \textbf{mAP}. 

As shown in Figure~\ref{fig:fitted_map_acc_curve}, the similar trend is shared between the curves of \textbf{mAP} and \textbf{ACC}. These indexes all exhibit rapid growth in first 12 epochs and improve slower in the subsequent epochs. From the slope (first-order derivative) curve of approximated polynomial, we can identify the end epoch of the early-learning phase represented as $\textbf{EL}$ (i.e. the initial epoch of memorization phase) by the condition that the second-derivative at $\textbf{EL}$ is approximately equal to 0. The condition is formulated as \eqref{equ:el_factor}:             

\begin{equation}
\left|\textbf{Poly}^{''}_{[\textbf{ACC}_1:\textbf{ACC}_{\textbf{EL}}]}(\textbf{EL})\right| < \eta  
\label{equ:el_factor}
\end{equation}
where $\textbf{Poly}^{''}_{[\textbf{ACC}_1:\textbf{ACC}_{\textbf{EL}}]}(\textbf{EL})$ is the second derivative of the polynomial at $\textbf{EL}$, and the polynomial is acquired by fitting the discrete $\textbf{ACC}$ values $[\textbf{ACC}_1:\textbf{ACC}_{\textbf{EL}}]$, and $\textbf{ACC}_{i}$ stands for the \textbf{ACC} value for $i$-th epoch. $\eta$ is a threshold and we set to 0.001 in experiments.

\subsection{Dynamic Loss Decay for Robust ORSOD}
\label{sssec:DLD}
\quad Different from ELR~\cite{liu2020early} which adds early-learning regularization term in loss function and ADELE~\cite{liu2022adaptive} which corrects high confidence false labels in early-learning stage, we design a more intuitive and effective scheme called dynamic loss decay to mitigate the influence of wrong category annotations in ORSOD. As Section \ref{sec:Pre}, in the early-learning stage, the optimization is dominated by the outnumbered correct labels and will be affected by noisy labels in memorization stage. Therefore, the identified \textbf{EL} indicates the right time of intervention to avoid the detrimental influence of noisy labels in loss back propagation and misleading the ORSOD model. 

The DLD mechanism is illustrated in Figure~\ref{fig:overview} and consists of two phases divided by \textbf{EL}. In the early-learning stage, i.e. the training epoch is smaller than \textbf{EL},  we use the standard Cross Entropy loss. In the memorization stage, we select the top K samples of which the losses are the top K largest ones (more probably noisy labels), and epoch-wisely decay the losses of the top K samples as they are most probably calculated with wrong category labels. All the training samples are denoted by $\bm{X}$, the top K samples with top K largest losses are represented as $\bm{X}_K$, and $\bm{X}_r$ stands for the rest samples. The formulation of $\mathcal{L}_\text{DLD}$ can be given in \eqref{eq:dld}: 

\begin{equation}
\mathcal{L}_\text{DLD} = 
\left\{
    \begin{array}{lr}
        \mathcal{L}_{\text{CE}}(\bm{X}), & \text{if } \text{EC} < \textbf{EL}\\
        \alpha \mathcal{L}_{\text{CE}}(\bm{X}_k)+\mathcal{L}_{\text{CE}}(\bm{X}_r), & \text{if } \text{EC} \geq \textbf{EL}
    \end{array} \right.
    \label{eq:dld}
\end{equation}

where $\alpha = \exp (\frac{10}{\text{EC}-\textbf{EL}})$ is the dynamic decay factor, $\mathcal{L}_{\text{CE}}(\bm{X})$ represents the Cross Entropy loss with respect to samples $\bm{X}$, EC stands for the current epoch number. In the memorization stage, the loss function consists of two parts: the first part gradually diminishes as the number of epochs increases, and the second part remains unchanged.

\begin{table}[ht]
\centering
\scalebox{0.9}{\begin{tabular}{c|c|c|c|c}
\multirow{2}{*}{\textbf{EL}} & 20\%-LSK & 30\%-LSK & 40\%-LSK & 20\%-LSK\\
 & Tiny(\textbf{12})  &  Tiny(\textbf{12})  & Tiny(\textbf{12}) & Small(\textbf{14}) \\
\hline
baseline & 70.5 & 70.3 & 69.0 & 73.2 \\ \hline
\textbf{EL}-8 & 70.7 & 70.8 & 70.3 & 72.8 \\
\textbf{EL}-4 & 71.2 & \textbf{72.3} & 69.1 &  72.6 \\
\hline
\textbf{EL} & \textbf{71.9} & 71.6 & \textbf{70.4} & \textbf{73.4} \\
\hline
\textbf{EL}+4 & 71.4 &  71.6 & 69.5 & 73.0 \\
\textbf{EL}+8 & 71.5 & 71.5 & 69.4 & 73.4 \\
\end{tabular}
}
 \caption{Ablation study of early-learning end point EL. We report the best \textbf{mAP} values acquired by varying the beginning epoch of the second stage of DLD on validation sets of DOTA-v1.0. The first row term ``20\%-LSK Tiny (12)" represents that the Oriented R-CNN model with LSK-Tiny backbone trained on DOTA-v1.0 containing 20\% category label noises, and the \textbf{EL} identified by our method is 12. We can observe that the identified $\textbf{EL}$ leads to superior robustness on different datasets and backbones.}
\label{tab:indicator_el}
\end{table}

\begin{table*}[!htbp]
\centering
\begin{tabular}{l|c|c|c|c|c|c|c|c}
\multirow{2}{*}{Dataset} & \multirow{2}{*}{Detector} & 0\% noise &\multicolumn{2}{c|}{20\% noise} &\multicolumn{2}{c|}{30\% noise} &\multicolumn{2}{c}{40\% noise}\\
\multirow{2}{*}{}  &  & baseline  & baseline  & \textbf{DLD}  & baseline  & \textbf{DLD} & baseline  & \textbf{DLD} \\
\bottomrule %[2pt]  
\multirow{3}{*}{DOTA-v1.0}    & Oriented R-CNN      & 74.2   & 70.5 & \textbf{71.9(+1.4)} & 70.3 & \textbf{71.6(+1.3)} & 69.0 & \textbf{70.4(+1.4)}   \\
\multirow{3}{*}{}            & ROI-Transformer      & 74.6 & 71.5 & \textbf{71.6(+0.1)} & 70.4 & \textbf{71.6(+1.2)} & 68.1 & \textbf{70.1(+2.0)}       \\
\multirow{3}{*}{}            & ReDet      & 73.2 & 70.5 & \textbf{70.5(+0.0)} & 69.9 & \textbf{70.3(+0.4)} & 67.7 & \textbf{68.3(+0.6)}       \\
\end{tabular}
 \caption{Comparison of \textbf{mAP}(\%) for different ORSOD models training  with DLD. The results of DLD are highlighted. We report the best \textbf{mAP} value for validation set of DOTA-v1.0.}
\label{tab:frameworks}
\end{table*}

\begin{table*}[!ht]
\centering
\begin{tabular}{c |c|c|c|c|c}
Method   & baseline & Top-K=3\% & Top-K=5\% & Top-K=7\% & Top-K=10\% \\
\hline
LSK-T-20\%    & 70.5       & 71.5     &  \textbf{71.9}   & 70.5   & 70.8      \\
LSK-T-30\%    & 70.3       & 71.0     &  \textbf{71.6}   & 71.0   & 70.9      \\
LSK-T-40\%    & 69.0       & 68.6    &  68.9   & \textbf{70.4}   & 70.1  \\
\end{tabular}
 \caption{Comparison of \textbf{mAP}(\%) for the hyperparameter \textbf{Top-K} with different proportions of category incorrect labels in the DOTA-v1.0 dataset. The selected model is the Oriented R-CNN detector with LSKNet-Tiny backbone. }
\label{tab:top-k}
\end{table*}

\section{Experiments}
\quad In this section, we first briefly introduce the testing data sets HRSC2016~\cite{Liu2017AHR}, DOTA-v1.0/v2.0~\cite{Xia_2018_CVPR,9560031}. Then, we describe the implementation details of the experiments. Thirdly, we report the ablation study of the key components and in-depth analysis of proposed method to verify their effectiveness. Finally, we show the performance of proposed method compared with other competitors on both synthesized noisy ORSOD dataset and 2023 National Big Data and Computing Intelligence Challenge.

\label{sec:experiments}

\subsection{Datasets}

\quad \textbf{HRSC2016}~\cite{Liu2017AHR} is a popular ship detection dataset that contains 1,070 images and 2,976 instances using satellite imagery. It has a three level category hierarchy, and we chose to use its first and second tier level which contains four categories: aircraft carrier, warship, merchant ship and other generic ships.

\textbf{DOTA-v1.0}~\cite{Xia_2018_CVPR} is a large scale aerial images dataset for object detection. It is widely used in develop and evaluate ORSOD methods. The dataset contains 15 categories, 2,806 images and more than 180,000 instances. The size of images varies from $800 \times 800$ to $20000\times 20000$. 

\textbf{DOTA-v2.0}~\cite{9560031} collects more sub-meter remote sensing and aerial images. DOTA-v2.0 has 18 categories, 11,268 images and 1,793,658 instances. In our ablation study, we have assessed 17 out of 18 categories in the validation set of DOTA-v2.0, excluding the "helipad" category. The validation set comprises a total of 130,909 instances, with only 3 instances belonging to the "helipad" category in our processed dataset. The rare presence of "helipad" instances significantly degrades the $\textbf{mAP}$ and $\textbf{ACC}$ indexes and distorts the overall curve trends. 

% All the images from each dataset were resized to $1024 \times 1024$ pixels as input. We used validation set for evaluating the results.

% \textbf{Introduction of noise}. For adding noise in categories. 

\subsection{Implementation}
\quad We adopt single-scale training and testing strategy by cropping all images into 1024 $\times$ 1024 patches with overlap of 200 pixels. For noisy category label generation, we randomly select a proportion of instances in the annotations, and set their categories to random new ones without changing the bounding box. Oriented R-CNN~\cite{xie2021oriented} on MMRotate~\cite{Zhou_2022} framework, with LSKNet~\cite{Li_2023_ICCV}-Tiny, LSKNet-Small and SwinTransformer~\cite{liu2021swin}-Tiny as backbones, is adopted for different experiments respectively. NVIDIA A40 GPU is utilized to carry out all the experiments. The models have been trained for 36 epochs with AdamW optimizer. We first train the baseline model on the original dataset with clean labels. Then, we train ORSOD models with different level of noisy labels (20\%, 30\%, 40\%) for comparisons. 

\subsection{Ablation Study and Analysis}

\quad In this section, we report the results of ablation study on validation sets of DOTA-v1.0 and DOTA-v2.0 to validate the effectiveness of our method. 

\textbf{End Point of Early-learning.} Identifying the endpoint of early learning is a crucial component of our method. We adhere to the approach outlined in Section \ref{sssec:early-learning} and compute \textbf{EL} using Equation \eqref{equ:el_factor}. The influence of changing the initial epoch of the second stage of DLD is explored in Table~\ref{tab:indicator_el}. The candidate epochs $[\textbf{EL}-8, \textbf{EL}-4, \textbf{EL}+4, \textbf{EL}+8]$ are selected around the \textbf{EL} found by our proposed criterion. In order to verify the generality of the identified \textbf{EL}, we report the best DOTA-v1.0 validation set \textbf{mAP} of different backbones (LSK-Tiny and LSK-Small) training with different noisy-level labels (20\%, 30\%, and 40\%). For example, the head row term ``20\%-LSK Tiny (12)" represents the experimental setting, that the Oriented R-CNN model with LSK-Tiny backbone trained on DOTA-v1.0 containing 20\% category label noises with DLD, and the corresponding \textbf{EL} identified by our method is 12. From the third row to the seventh row of the second column, the initial epoch of loss decay in DLD is varied from 4 (EL-8) to 20 (EL+8). 

We can observe that the $\textbf{EL}$ selected by our proposed criterion consistently lead to superior robustness with different noise-level datasets and backbones. Although the endpoint \textbf{EL} identified by the method in Section \ref{sssec:early-learning} serves as a effective indicator for DLD, it's important to note that the best performance may not achieved exactly at \textbf{EL}, but in the neighbourhood of \textbf{EL} as shown in the third column of Table~\ref{tab:indicator_el}. Therefore, comparing with the baseline model,  we can claim that under the guidance of \textbf{EL}, DLD can significantly boost its the noise resistance. 

\begin{figure}[ht]
\centering
\includegraphics[width=0.9\linewidth]{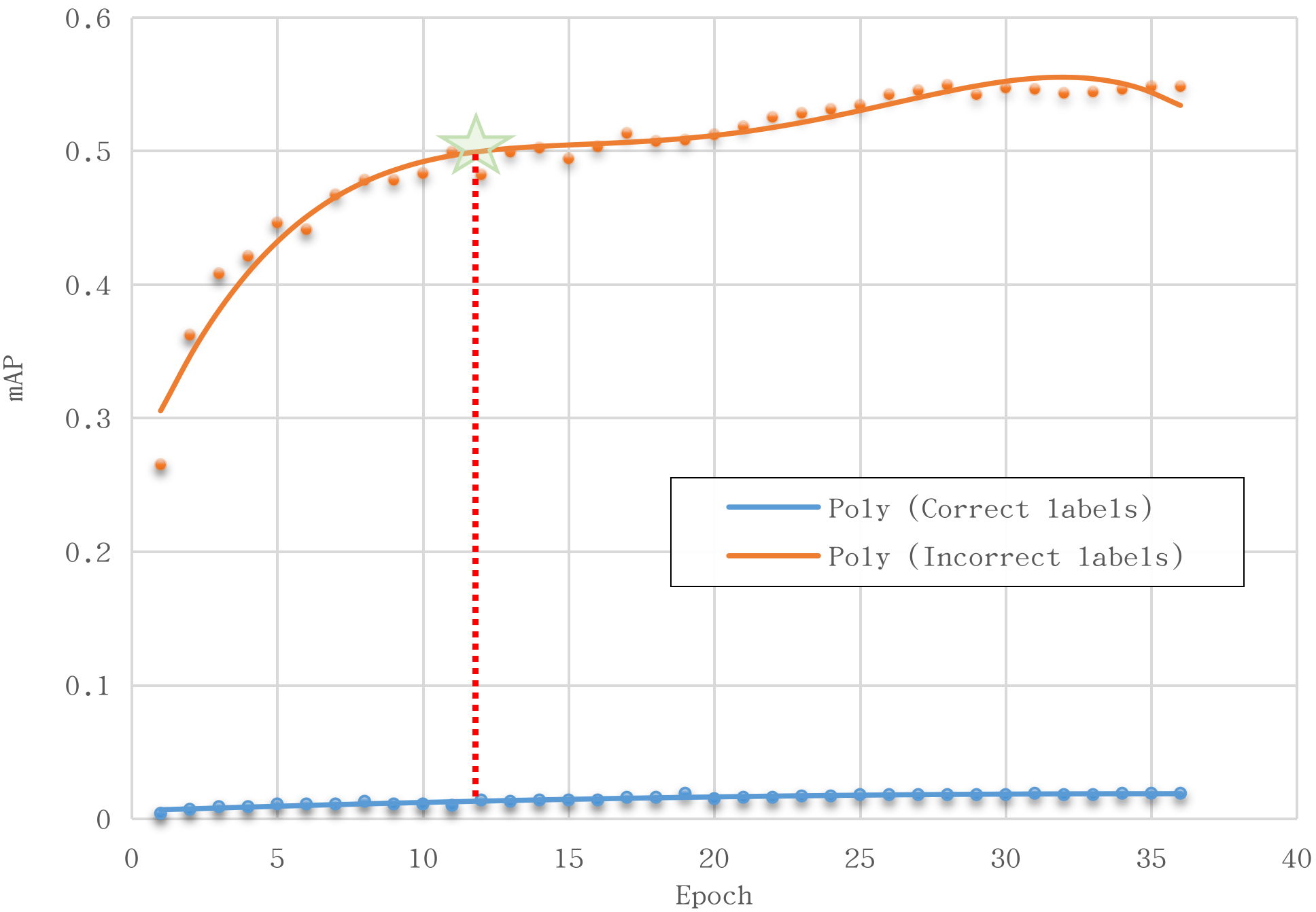}
\caption{ The more elaborated curves of \textbf{mAPC} (\textbf{mAP} with respect to correct category labels) and \textbf{mAPI} (\textbf{mAP} with respect to incorrect category labels) of training set. The Orient R-CNN model is trained with labels with 40\% noise level. The curve of \textbf{mAPI} stays below 2\% while \textbf{mAPC} continuously improves during whole process. }
\label{fig:mAPC_mAPI}
\end{figure}

To further analysis the relationship between \textbf{EL} and mAP accuracy of incorrect labels, we show the more elaborated curves of \textbf{mAPC} (\textbf{mAP} with respect to correct category labels) and \textbf{mAPI} (\textbf{mAP} with respect to incorrect category labels) in training set in Figure \ref{fig:mAPC_mAPI}. The  ORSOD baseline model is trained using labels with 40\% noise.  \textbf{mAPC} continuously improves, and slow down its growth rate at epoch 12 which is align with \textbf{mAP} performance in validation set as is shown in Figure \ref{fig:fitted_map_acc_curve}. This is phenomena also demonstrates that \textbf{ACC} can reflect the overall training accuracy of correct labels when model train with noisy labels.  To our surprise, \textbf{mAPI} stays below 2\% without a sharp increase during whole process.

\textbf{ The effectiveness of DLD.} We demonstrate that the effectiveness of DLD from three aspects: analysis \textbf{ACC} curves, class activation map visualization and integration with different ORSOD methods. 

\begin{figure}[ht]
\centering
\includegraphics[width=0.9\linewidth]{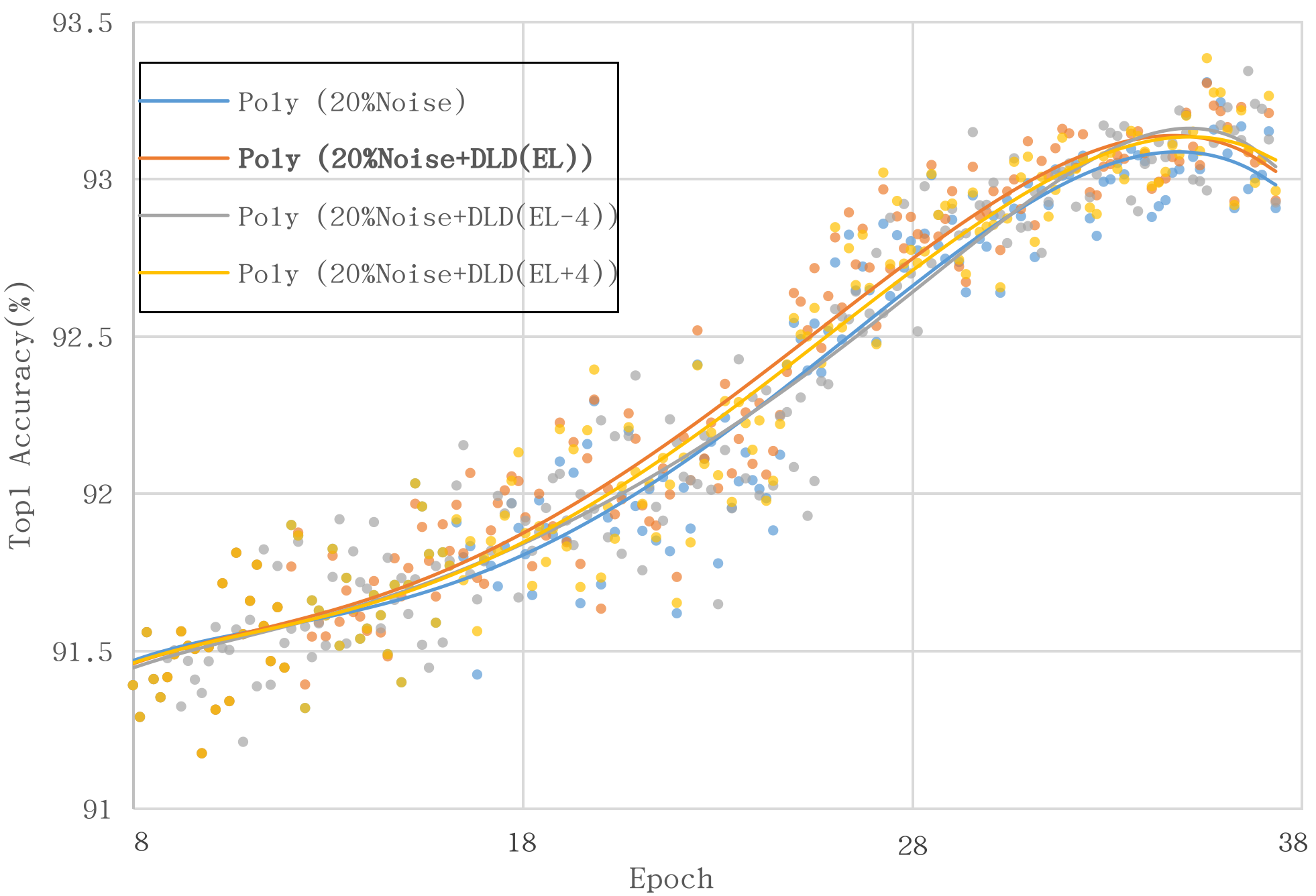}
\caption{ The \textbf{ACC} curves of ORSOD model training with four different strategies and labels with 20\% noise. The four training strategies are training Oriented R-CNN not using DLD, using DLD and the loss decay begins at epoch EL-4 (8), EL (12), and EL+4 (16).}
\label{fig:acc_dld}
\end{figure}

Firstly,  we compare the \textbf{ACC} curves of ORSOD model training with four different strategies and labels with 20\% noise in Figure \ref{fig:acc_dld}. The four training strategies are not using DLD, using DLD and the loss decay begins at epoch \textbf{EL}-4 (8), \textbf{EL} (12), and \textbf{EL}+4 (16). The three training strategies with DLD consistently improve the \textbf{ACC}, which shows that DLD take effects at relatively relaxed \textbf{EL} neighbourhood. 

\begin{figure*}[ht]
\centering
\includegraphics[width=1\linewidth]{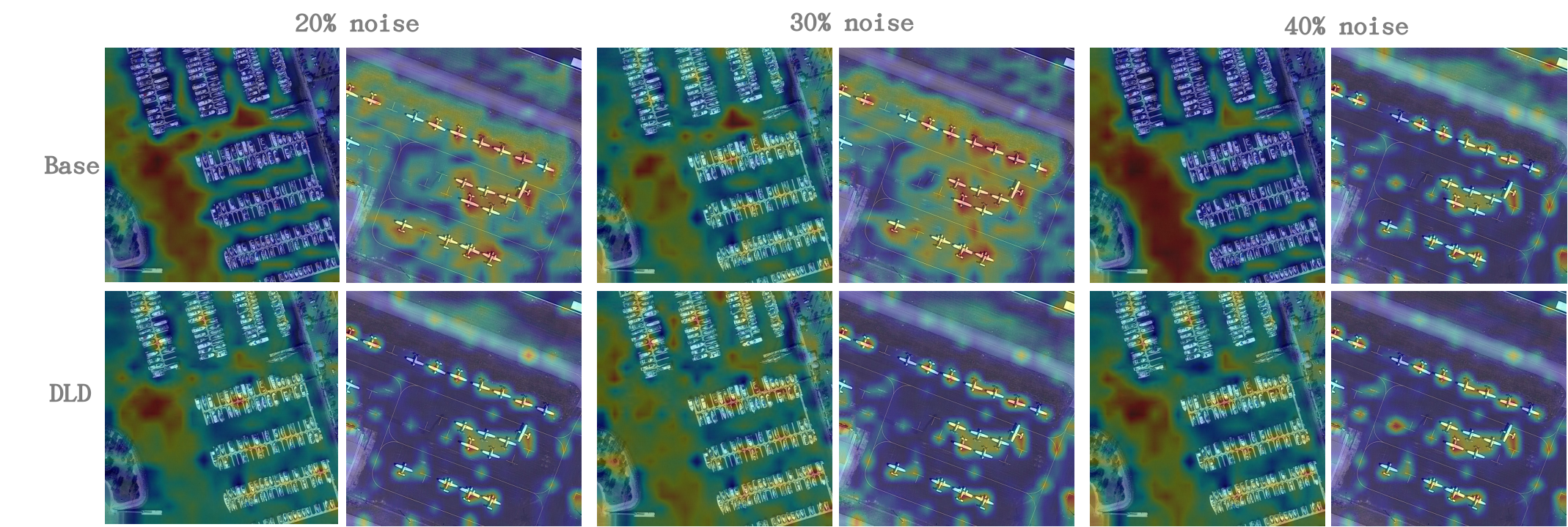}
\caption{The heatmap shows the distribution of region of interest with the attention map during inference, where the attention increases from blue to red. Notably, the ORSOD model is more focused on the target objects after DLD is applied.}
\label{fig:heatmap}
\end{figure*}

Secondly, we compare the class activation maps of ORSOD model with and without DLD on test images. Some examples of EigenCAM \cite{9206626}  are shown in  Figure \ref{fig:heatmap}. Training with DLD, the ORSOD model tend to focus more on foreground targets which is beneficial for noise resistance. 

Finally, to further analysis the generality of our method for different oriented object detectors, we choose two representative methods ROI-Transformer and ReDet \cite{han2021redet}, and employ LSKNet-Tiny as backbone. Meanwhile, we use DOTA-v1.0 dataset with different proportion of noise for training. The results are illustrated in Table \ref{tab:frameworks} which demonstrates that our method can be easily integrated in different ORSOD models and improve their robustness.

% For better visualize detection results examples of EigenCAM \cite{9206626} shown in  Fiugre \ref{fig:heatmap}. Training with DLD, attention of model tend to focus more on foreground targets, leading to boost in mAP accuracy.

\textbf{Top-K Selection.} The effectiveness of DLD heavily relies on the hyperparameter K in selecting the top K samples with largest losses. We have conduct extensive experiments to explore the impact of different K values, specifically top 3\%, top 5\%, top 7\%, and top 10\%. As illustrated in Table ~\ref{tab:top-k}, when the incorrect label proportions are 20\% and 30\%, top 5\% yields the highest \textbf{mAP} , while top 7\% performs better in the case of 40\% noise. These experimental findings conform with the intuitive expectation that dataset with larger proportion of incorrect labels should employ larger K value to decay the weight of more losses terms.

\begin{table*}[!h]
\centering
\scalebox{0.97}{\begin{tabular}{c|c|c|c|c|c|c|c|c}
\multirow{2}{*}{Dataset} & \multirow{2}{*}{Method} & 0\% noise &\multicolumn{2}{c|}{20\% noise} &\multicolumn{2}{c|}{30\% noise} &\multicolumn{2}{c}{40\% noise}\\
\multirow{2}{*}{}  &  & baseline  & baseline  & \textbf{DLD}  & baseline  & \textbf{DLD} & baseline  & \textbf{DLD} \\
\bottomrule %[2pt]  
\multirow{3}{*}{DOTA-v1.0}    & LSKNet-Tiny      & 74.2   & 70.5 & \textbf{71.9(+1.4)} & 70.3 & \textbf{71.6(+1.3)} & 69.0 & \textbf{70.4(+1.4)}   \\
\multirow{3}{*}{}            & LSKNet-Small      & 75.6 & 73.2 & \textbf{73.4(+0.2)} & 70.7 & \textbf{72.2(+1.5)} & 70.2 & \textbf{71.0(+0.8)}       \\
\multirow{3}{*}{}            & SwinTransformer-Tiny      & 75.3 & 71.5 & \textbf{71.7(+0.2)} & 71.4 & \textbf{72.0(+0.6)} & 69.5 & \textbf{70.8(+1.3)}       \\
\bottomrule %[2pt] 
\multirow{3}{*}{DOTA-v2.0}    & LSKNet-Tiny      & 66.1 & 63.5 & \textbf{61.5(+1.4)} & 63.1 & \textbf{63.7(+0.6)} & 62.4 & \textbf{62.8(+0.4)}       \\
\multirow{3}{*}{}            & LSKNet-Small     & 66.4 & 63.1 & \textbf{64.2(+1.1)} & 62.9 & \textbf{64.3(+1.4)} & 62.2 & \textbf{63.2(+1.0)}       \\
\multirow{3}{*}{}            & SwinTransformer-Tiny      & 67.2 & 63.8 & \textbf{64.4(+0.6)} & 62.5 & \textbf{62.9(+0.4)} & 61.3 & \textbf{62.8(+1.5)}       \\
\bottomrule %[2pt] 
\multirow{3}{*}{HRSC2016}    & LSKNet-Tiny       & 82.1 & 81.1 & \textbf{81.3(+0.2)} & 78.6 & \textbf{79.2(+0.6)} & 72.7 & \textbf{72.9(+0.2)}       \\
\multirow{3}{*}{}            & LSKNet-Small      & 86.8 & 85.3 & \textbf{85.6(+0.3)} & 83.0 & \textbf{84.7(+1.7)} & 80.1 & \textbf{81.7(+1.6)}       \\
\multirow{3}{*}{}            & SwinTransformer-Tiny    & 75.3 & 71.5 & \textbf{71.7(+0.2)} & 71.4 & \textbf{72.0(+0.6)} & 69.5 & \textbf{70.8(+1.3)}        \\
\end{tabular}
}
 \caption{Comparison between baseline and DLD method on three datasets: DOTA-v1.0, DOTA-v2.0, and HRSC2016. These datasets have been deliberately contaminated with category noise levels of 20\%, 30\%, and 40\%. The employed detector employed is Oriented R-CNN, and three distinct backbones are selected: LSKNet-Tiny, LSKNet-Small, and SwinTransformer-Tiny. The reported values represent the highest accuracy from the best epoch evaluated in the validation set. DLD represents the model that is trained by our method.}
 \label{tab:ablation study}
\end{table*}

\subsection{Final Results}

\quad \textbf{Test on Open Datasets.} Based on Oriented R-CNN framework, we evaluate DLD with backbones of three different size (CNN based LSKNet-Tiny : 4.3M\cite{Li_2023_ICCV}, CNN-based LSKNet-Small : 14.4M\cite{Li_2023_ICCV}, and Transformer based SwinTransformer-Tiny: 29M\cite{liu2021swin}), three ORSOD datasets (DOTA-v1.0, DOTA-v2.0, and HRSC2016), three proportion of incorrect category labels(20\%, 30\%, and 40\%). The results presented in Table \ref{tab:ablation study} demonstrates that DLD can effectively alleviate the models' degradation caused by noisy category labels, the maximum improvement in \textbf{mAP} is 2.0\% for ROI-Transformer. Meanwhile, as is shown in Table \ref{equ:el_factor}, endpoint of early-learning for Oriented Object Detectors can be identified effectively through our method.

\textbf{DLD vs. Label Smoothing Method~\cite{krizhevsky2012regularizing}.} Label Smoothing (LS) is a regularization method involves penalizing the distribution of the network's outputs, thereby encouraging the model to be more cautious in its predictions. As shown in Table \ref{tab:soft label}, DLD outperforms LS by a considerable margin in the experiments with 20\% and 30\% category incorrect labels, and achieves an equal mAP in the 40\% noise ratio setting.
\begin{table}[!htbp]
\centering
\begin{tabular}{c|c|c|c}
Method   & baseline & LS & \textbf{DLD}  \\
\hline
LSK-T-20\%    & 70.5       & 71.5     &  \textbf{71.9}         \\
LSK-T-30\%    & 70.3       & 70.9     &  \textbf{72.3}         \\
LSK-T-40\%    & 69.0       & \textbf{70.4}     &  \textbf{70.4}                  \\

\end{tabular}
 \caption{Comparison of mAP(\%) between Label Smoothing (LS) method and our method (DLD).}
\label{tab:soft label}
\end{table}

\textbf{NBDCIC 2023.} The competition comprise 7 sub-competitions and has attracted a total of 2,852 participating teams. Our team successfully implement the algorithm described above in the sub-competition "Fine-grained Object Detection Based on Sub-meter Remote Sensing Images". We have secured the second place among 383 competing teams, just 0.1\% away from the first-place winner. The competition dataset involves noisy categories annotations on sub-meter remote sensing images, encompasses 98 target categories of various types of aircraft to ships. The dataset is split into a training set, consisting of 8000 images and 101,054 instances, and a test set comprising 3000 images.

The competition has posed several significant challenges such as weak target features, subtle differences between classes, labeling noise, and an extremely unbalanced distribution of categories. The most noteworthy characteristic of the dataset is the category labels are randomly contaminated with noise and the ratio of label noise is different in the preliminary round and the final round. The competition pay close attention on the way of alleviating the degradation of model caused by category noise. Despite these challenges, our team has acquired remarkable results, as highlighted in Table \ref{tab:big data challenge} showcasing the superiority of proposed method in tackling label noise. 

\begin{table}[!ht]
\centering
\begin{tabular}{l|c|c}
\small
Team Name                     & Final Round & First Round \\
\hline
GaoKongTanCe Team    & 0.7758      & 0.7533        \\
\rowcolor{gray!40}Sensing\_earth(\textbf{ours}) & 0.7758      & 0.7518     \\
CAE\_AI             & 0.7565      & 0.6994        \\
Happy Children’s Day             & 0.7508      & 0.7656        \\
default13253462               & 0.7448      & 0.6803       
\end{tabular}
 \caption{Results of \textit{the Fine-grained Object Detection Based on Sub-meter Remote Sensing Images} from NBDCIC 2023.}
 \label{tab:big data challenge}
\end{table}

% \section{Discussion}

% \quad 1. Although the top k zone contains a lot of noisy losses, we can not just dump this part, because before a model is fully trained top k zone still contains quiet amount of losses for clean annotations. 

% 2. For extremely tiny dataset like HRSC2016, the accuracy of class head may drop hardly in the first several epochs.

% 3. Our investigation is based on remote sensing field, this method may have the potential in other fields but need further exploration. \\
\section{Conclusions and Future Works}
\quad In this paper, we propose the first robust oriented object detection method for remote sensing images, which addresses the issue of noisy category training labels with dynamic loss decay mechanism. Based on the theoretical analysis and experimental validation, we identify the existence of the early-learning phase and memorization phase in training ORSOD model with noisy labels, and propose a feasible approach to find the end point of early-learning $\textbf{EL}$. Accordingly, we design an effective dynamic loss decay scheme by gradually reduce the top K largest loss terms which are most likely calculated with false labels in subsequent epochs of $\textbf{EL}$. Experiments on both synthesized noisy ORSOD datasets and NBDCIC 2023 demonstrate the effectiveness of proposed DLD in preventing training category noise, and the ablation studies corroborate the rationality of the selected EL and the loss decay scheme. 

\quad For future work, our work mainly focus on the category noise in ORSOD, but negative effect caused by inaccurate OBB has been ignored, how to apply DLD to the combination of both category and OBB noise still need further exploration. Moreover, the $\alpha$ illustrated in the intuitive equation \eqref{eq:dld} plays an important role in the whole training process, we hope to find a better design to further improve the performance of DLD.

% \clearpage
{
    \small
    \bibliographystyle{ieeenat_fullname}
    \bibliography{main}
}
% \input{sec/appendix}
% \section{Rationale}
% \label{sec:rationale}
% % 
% Having the supplementary compiled together with the main paper means that:
% % 
% \begin{itemize}
% \item The supplementary can back-reference sections of the main paper, for example, we can refer to \cref{sec:intro};
% \item The main paper can forward reference sub-sections within the supplementary explicitly (e.g. referring to a particular experiment); 
% \item When submitted to arXiv, the supplementary will already included at the end of the paper.
% \end{itemize}

% To split the supplementary pages from the main paper, you can use \href{https://support.apple.com/en-ca/guide/preview/prvw11793/mac#:~:text=Delete%20a%20pccccccage%20from%20a,or%20choose%20Edit%20%3E%20Delete).}{Preview (on macOS)}, \href{https://www.adobe.com/acrobat/how-to/delete-pages-from-pdf.html#:~:text=Choose%20%E2%80%9CTools%E2%80%9D%20%3E%20%E2%80%9COrganize,or%20pages%20from%20the%20file.}{Adobe Acrobat} (on all OSs)
% , as well as \href{https://superuser.com/questions/517986/is-it-possible-to-delete-some-pages-of-a-pdf-document}{command line tools}.

\clearpage
\setcounter{page}{1}
\renewcommand\thesection{\Alph{section}}
\setcounter{section}{0}
\maketitlesupplementary

\section{Dynamic mAP results in validation dataset}
\label{sec:results in validation set}
\quad We have evaluated the mean Average Precision (mAP) of our model on the validation dataset, as presented in Table \ref{tab:ablation study}. This table encapsulates the peak mAP value attained throughout the entire training process. The fitted mAP curves for selected experiments are illustrated in Figure \ref{fig:val_el}, providing a visual representation of the overall changes and performance trends.

\textbf{Comparison between Baseline and DLD.} As illustrated in \textbf{Figure \ref{fig:val_el} (a), (c)}, and \textbf{(e)}, the impact of DLD on the endpoint \textbf{EL} is evident. Comparing the mAP curve after the application of DLD to the baseline, it is obvious that the DLD-enhanced curve consistently maintains a higher position. This result demonstrates that the mAP improvement achieved through DLD is not confined to a singular point but is discernible across the entire curve.

\textbf{Applying DLD at different endpoints.} As illustrated in \textbf{Figure 8 (b), (d)}, and \textbf{(f)}, we present visualizations of the mean average precision (mAP) fitted curves corresponding to different endpoints(\textbf{EL-8}, \textbf{EL-4}, \textbf{EL}, \textbf{EL+4} and \textbf{EL+8}). Results align with Table \ref{tab:indicator_el}. \textbf{EL} emerges as a robust indicator for the endpoint of early learning, meanwhile the experiment of training with 30\% noisy labels, \textbf{EL}-4 shows slightly better performance.

\section{Details of 2023 National Big Data and Computing Intelligence Challenge}
\label{sec:NBDCIC 2023}
\quad \textbf{Training data.} Dataset of this competition comprises a total of 98 categories encompassing ships and airplanes, as detailed in Table \ref{traning data of competiton}. The competition unfolds in two rounds. In the Initial Round, 10\% to 20\% of category labels are deliberately altered to incorrect ones. For the Final Round, an even higher proportion of labels undergo modification, with no specific information provided regarding the exact ratio.

\textbf{Our methods.} In this competition, three methods play a pivotal role in enhancing our model's performance: \textbf{DLD(ours)}, \textbf{A Bag of Tricks (ABT)}, and \textbf{Rotation Weighted Bbox Fusion (RWBF)}. Our method is based on framework of Oriented-Rcnn. Within ABT, we incorporate widely used data augmentation techniques such as multi-scale resizing, random rotation, and random flipping. Additionally, strategic training skills like Stochastic Weight Average (SWA) and Exponential Moving Average (EMA) are applied. We conduct a comparison of different backbones, including LSKNet-Small, Swin Transformer-Base, and InternImage-Base, to identify the optimal feature extractor. To amplify our model's performance even further, we integrate RWBF into the inference process. 

\textbf{Ablation results.} As depicted in Figure \ref{fig:competition}, these  experiments, same as the findings in Table \ref{tab:ablation study}, reveal that DLD consistently enhances the mAP accuracy of models across various backbones. In this competition, the maximum mAP improvement observed is \textbf{1.91\%} from InternImage-Base. Moreover, application of ABT and RWBF contributes significantly to our leaderboard score.

\section{Supplementary figures}
\label{sec:Supplementary figures}
\begin{figure*}[h]
    \centering
    \begin{subfigure}[b]{0.45\textwidth}
	\begin{minipage}[t]{\linewidth}
		\includegraphics[width=3in]{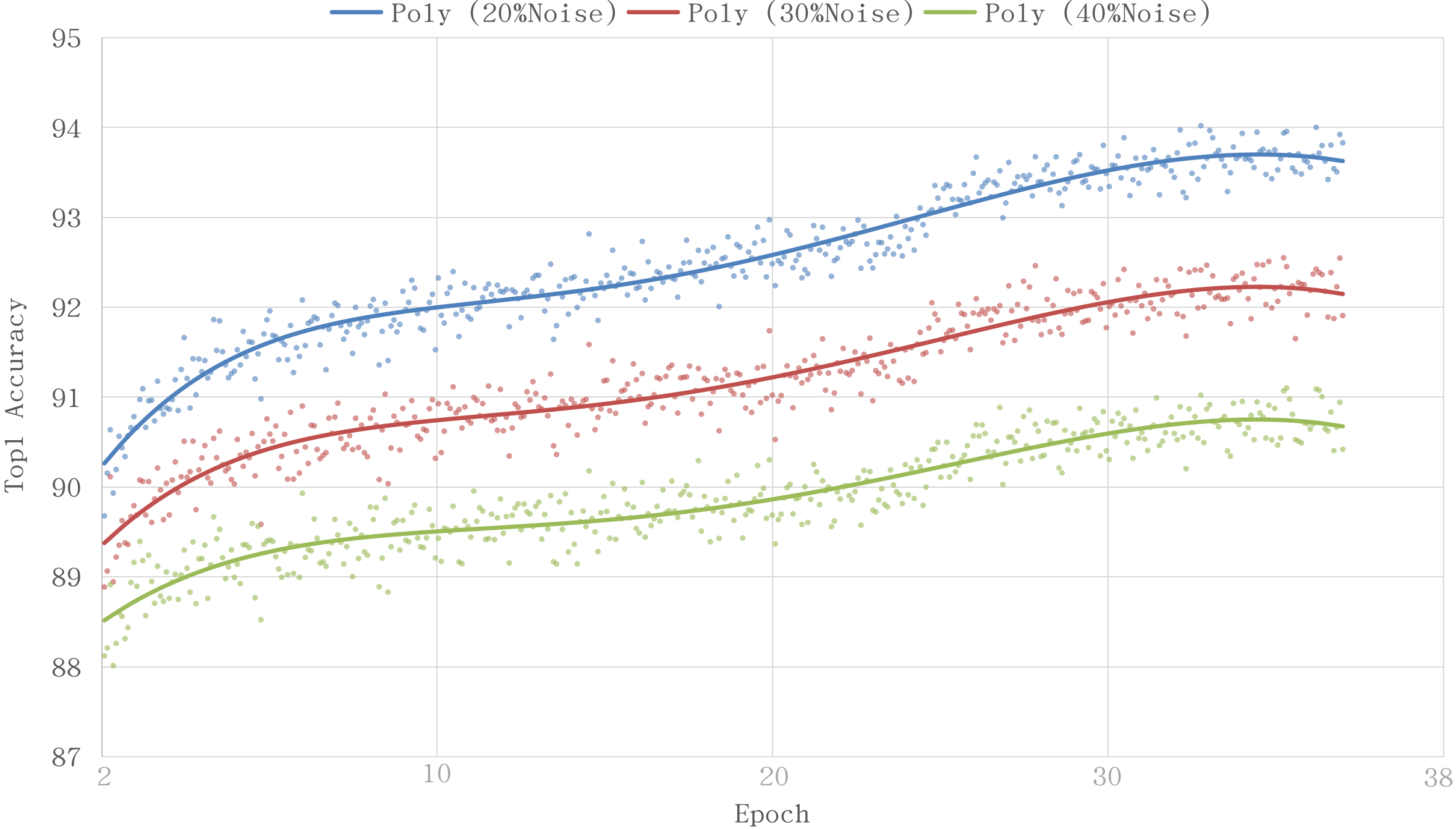} 
            \label{val_map}
            \caption{ \textbf{ACC-lsk-s}}
	\end{minipage}
    \end{subfigure}
    \begin{subfigure}[b]{0.45\textwidth}
	\begin{minipage}[t]{\linewidth}
		\includegraphics[width=3in]{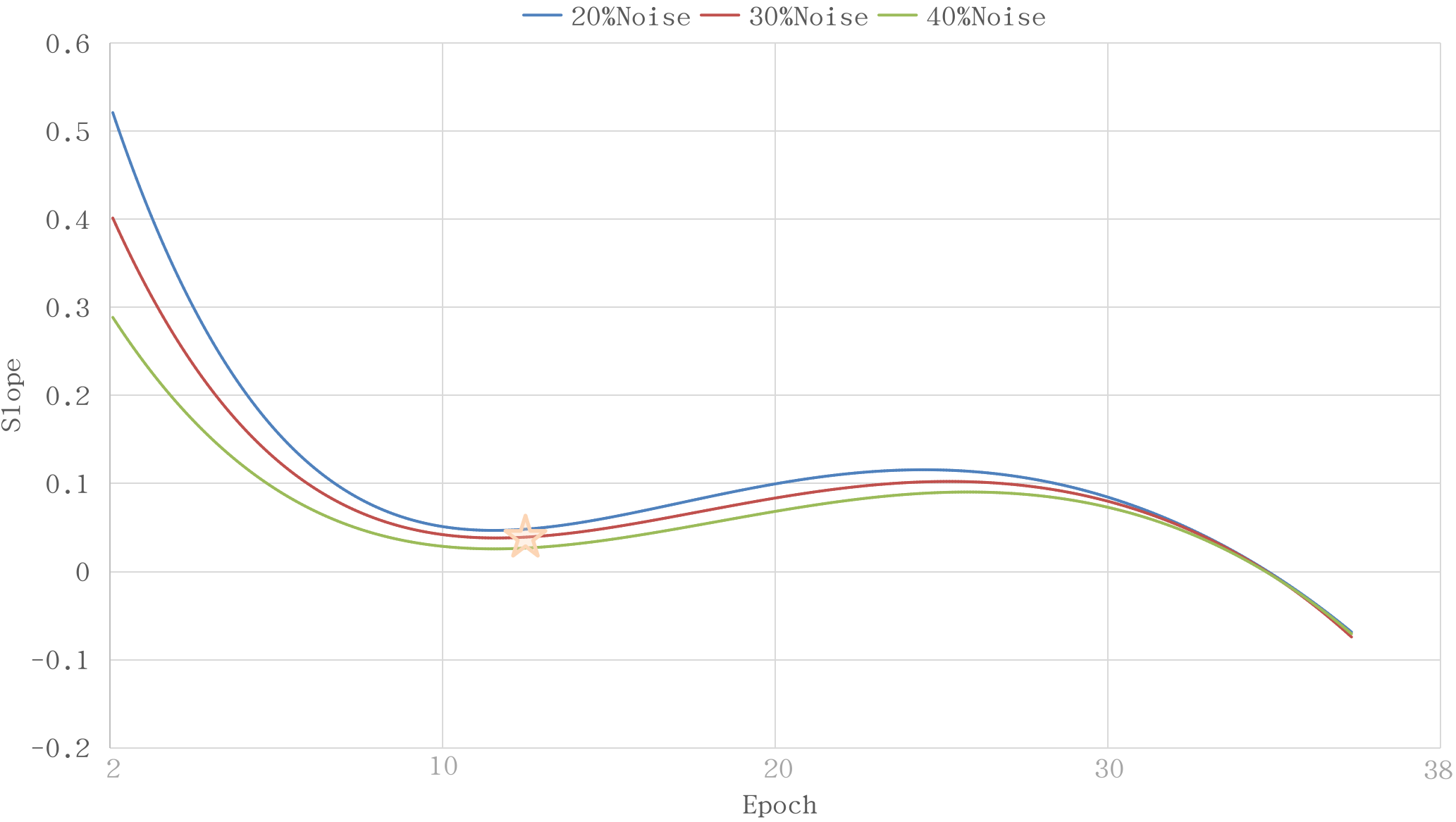}
            \label{val_map_slope}
            \caption{Slope of \textbf{ACC-lsk-s}}
	\end{minipage}
    \end{subfigure}
    \\ 
    \begin{subfigure}[b]{0.45\textwidth}
	\begin{minipage}[t]{\linewidth}
		\centering
		\includegraphics[width=3in]{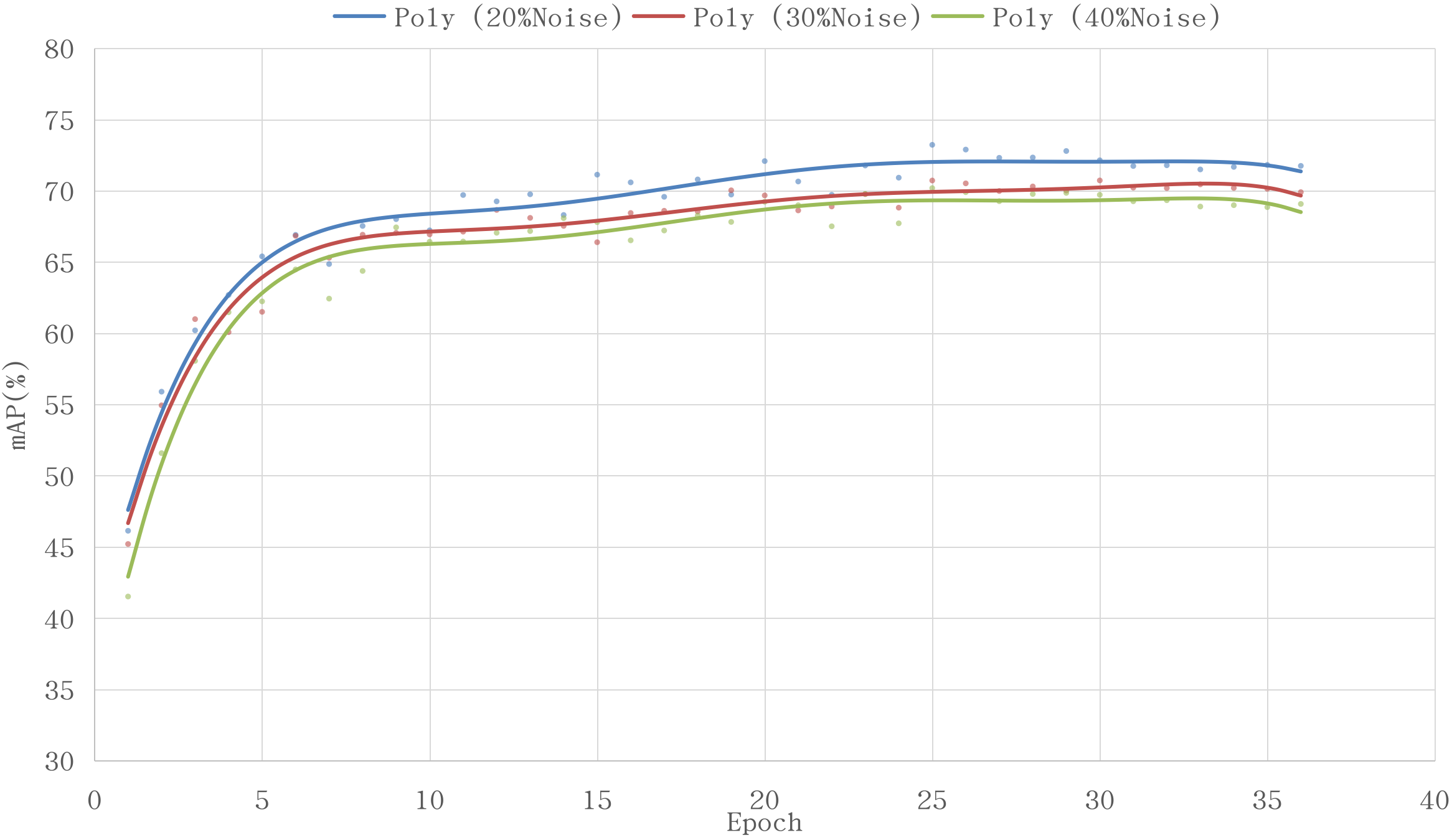}
            \label{fitted_acc}
            \caption{ \textbf{mAP-lsk-s}}
	\end{minipage}
    \end{subfigure}
    \begin{subfigure}[b]{0.45\textwidth}
	\begin{minipage}[t]{\linewidth}
		\centering
		\includegraphics[width=3in]{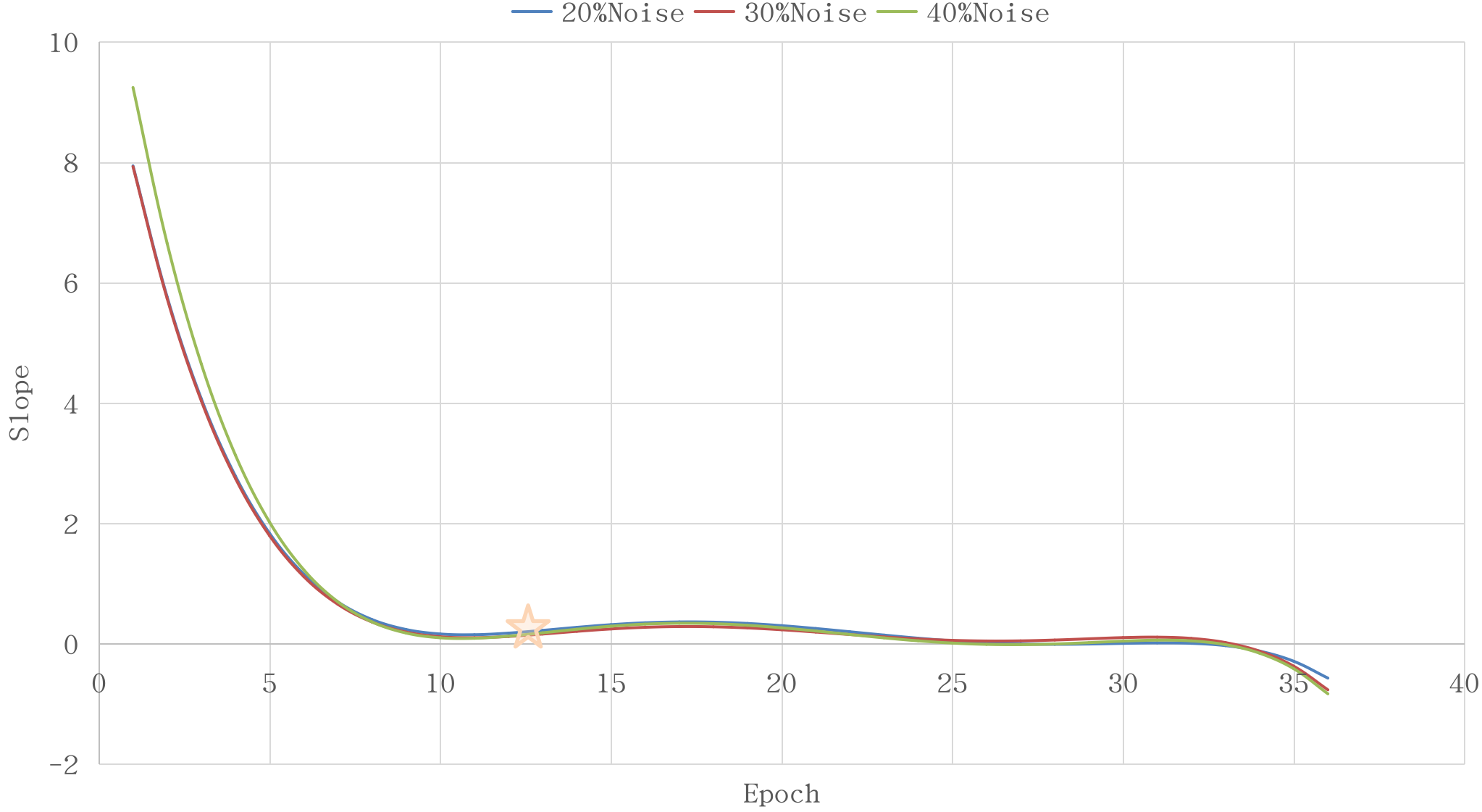}
            \label{acc_slope}
            \caption{Slope of \textbf{mAP-lsk-s}}
	\end{minipage}
    \end{subfigure}
    \\
        \begin{subfigure}[b]{0.45\textwidth}
	\begin{minipage}[t]{\linewidth}
		\includegraphics[width=3in]{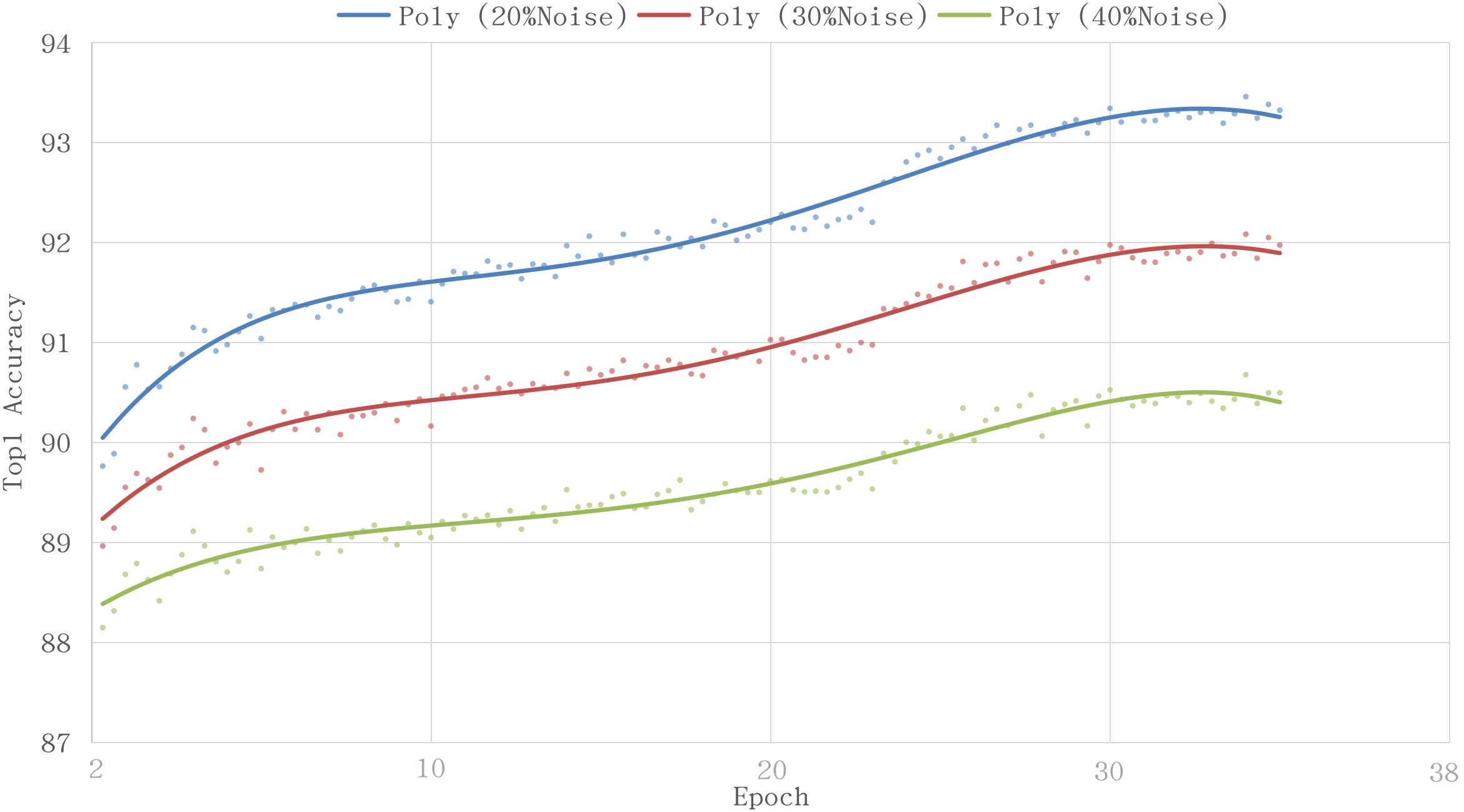} 
            \label{val_map}
            \caption{\textbf{ACC-swin-t}}
	\end{minipage}
    \end{subfigure}
    \begin{subfigure}[b]{0.45\textwidth}
	\begin{minipage}[t]{\linewidth}
		\includegraphics[width=3in]{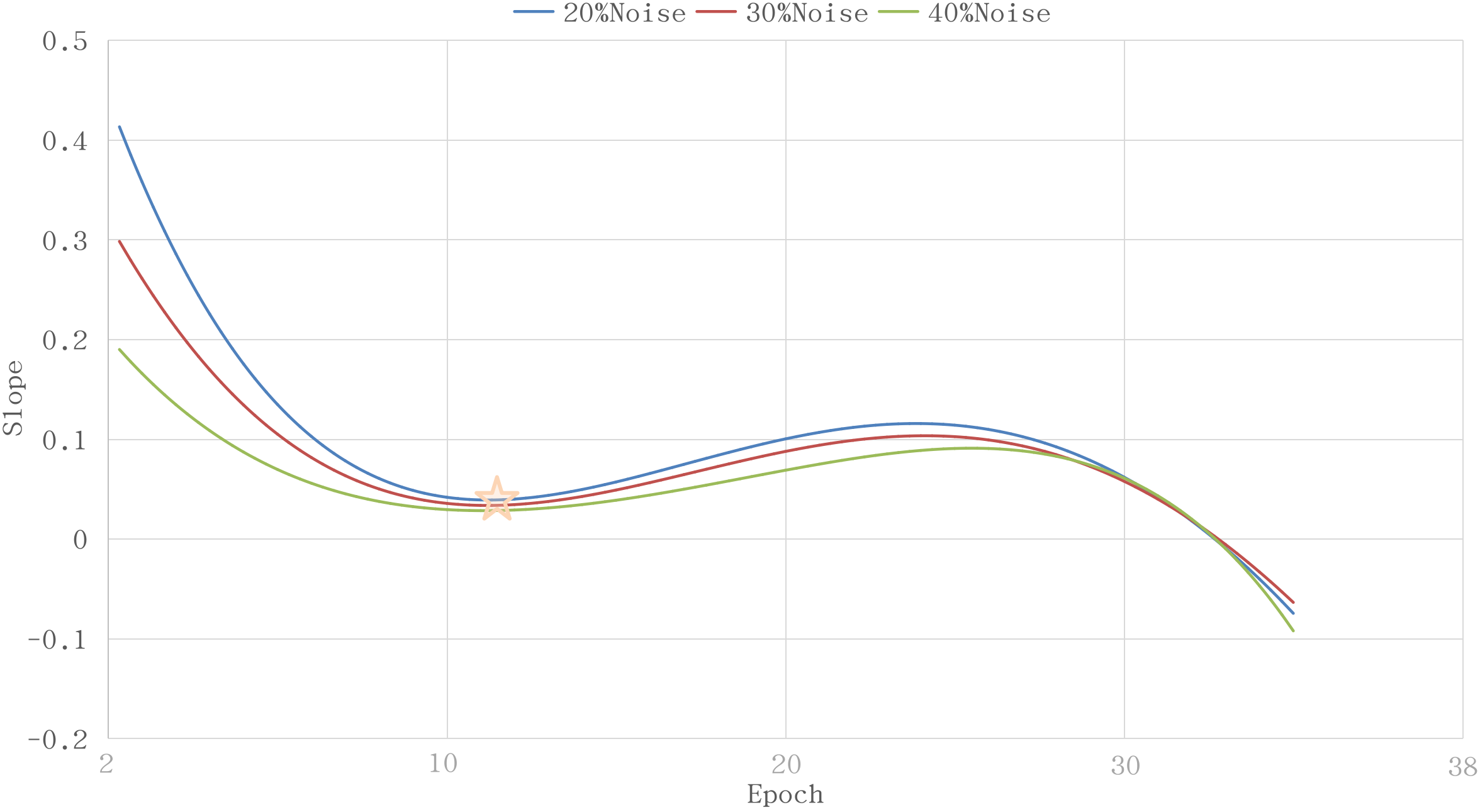}
            \label{val_map_slope}
            \caption{Slope of \textbf{ACC-swin-t}}
	\end{minipage}
    \end{subfigure}
    \\ 
    \begin{subfigure}[b]{0.45\textwidth}
	\begin{minipage}[t]{\linewidth}
		\centering
		\includegraphics[width=3in]{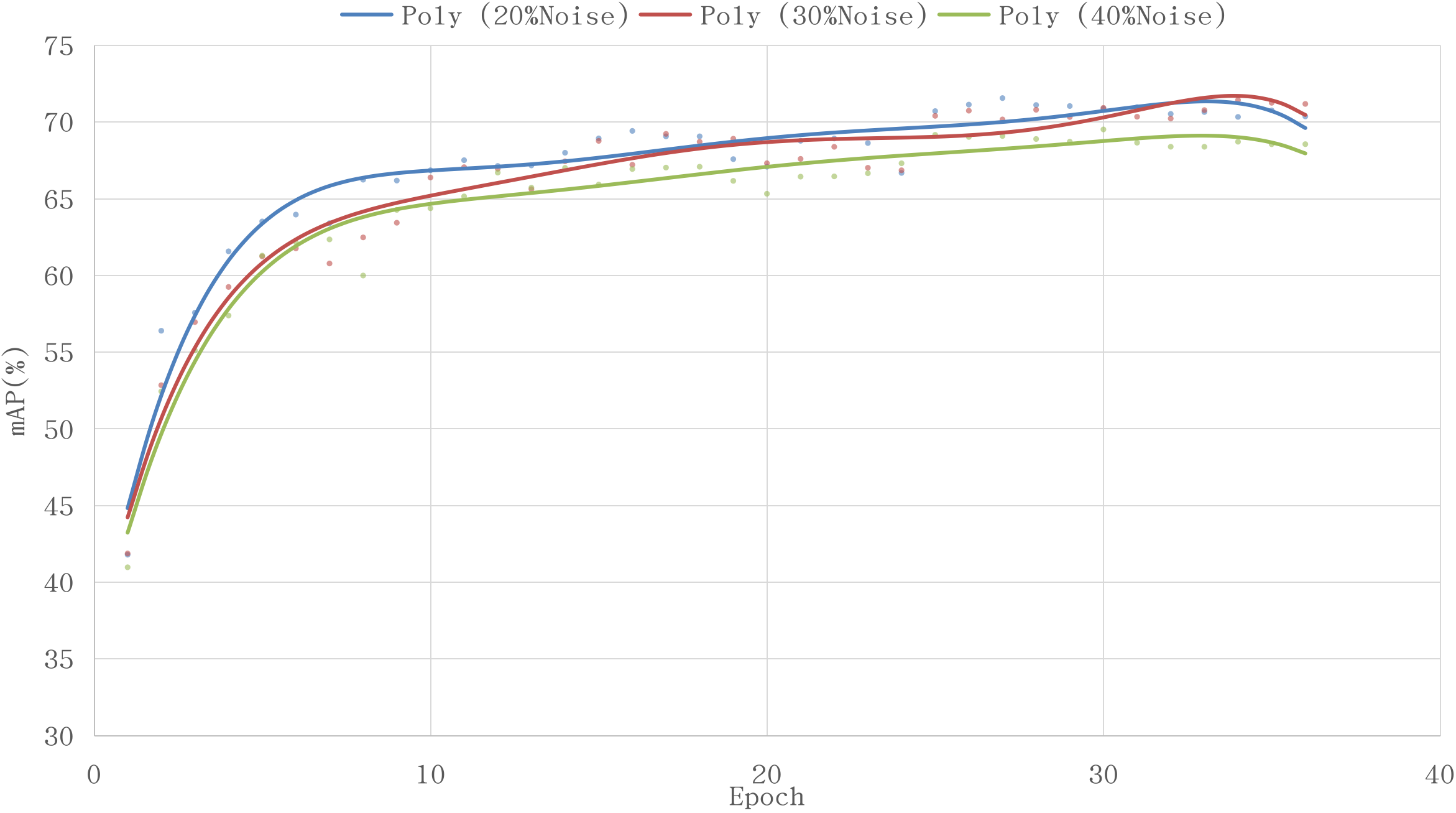}
            \label{fitted_acc}
            \caption{\textbf{mAP-swin-t}}
	\end{minipage}
    \end{subfigure}
    \begin{subfigure}[b]{0.45\textwidth}
	\begin{minipage}[t]{\linewidth}
		\centering
		\includegraphics[width=3in]{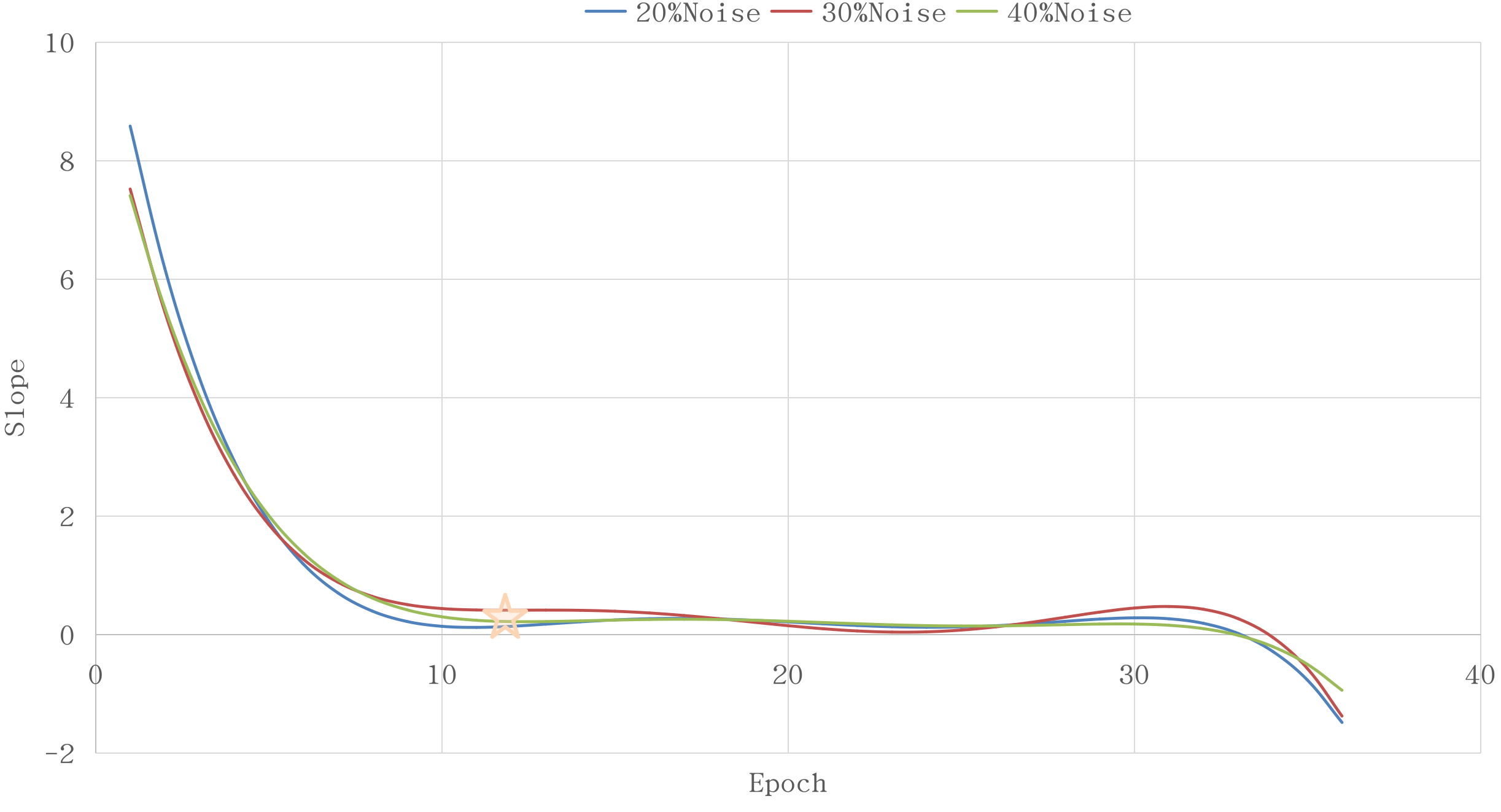}
            \label{acc_slope}
            \caption{Slope of \textbf{mAP-swin-t}}
	\end{minipage}
    \end{subfigure}
    \caption{The dynamics of two measurements mean average precision (\textbf{mAP}) and top 1 accuracy (\textbf{ACC}), acquired by the  Oriented R-CNN with \textbf{LSKNet-Small(lsk-s)} and \textbf{Swin Transformer-Tiny(swin-t)} backbone. The experiments are conducted on DOTA-v1.0 dataset contaminated with different level of category noises (20\%, 30\%, and 40\%). The \textbf{mAP} is calculated between model output and clean GT category labels. The \textbf{ACC} of the model output is referenced with noisy category labels. The star in (b)(d) represents the early-learning endpoint.}
    \label{fig:suppl_acc_map}
\end{figure*}

\begin{figure*}[h]
    \centering
    \begin{subfigure}[b]{0.45\textwidth}
	\begin{minipage}[t]{\linewidth}
		\includegraphics[width=3in]{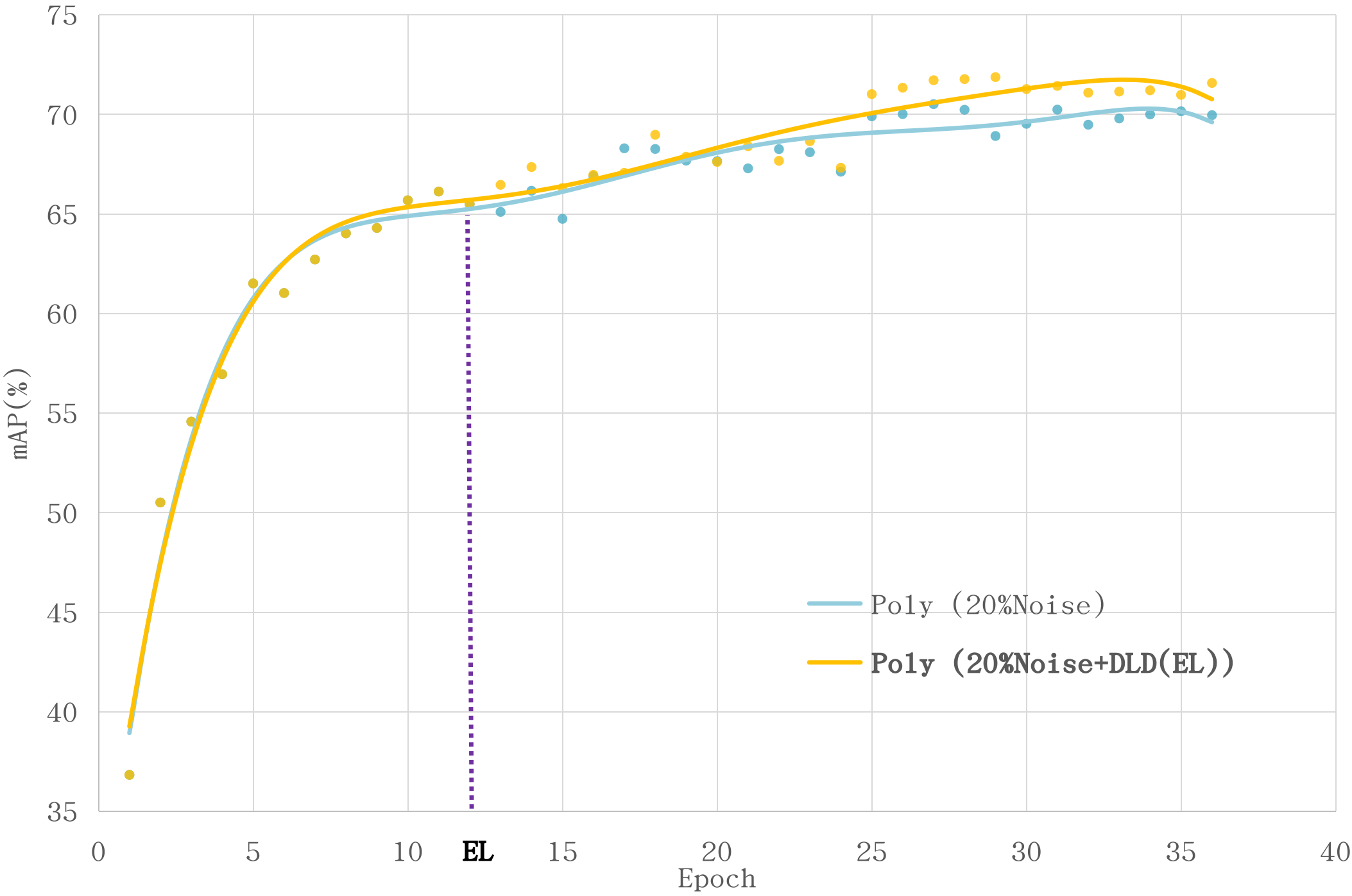} 
            \label{val_map}
            \caption{mAP\textbf{(Baseline vs. DLD)} 20\% Noise}
	\end{minipage}
    \end{subfigure}
    \begin{subfigure}[b]{0.45\textwidth}
	\begin{minipage}[t]{\linewidth}
		\includegraphics[width=3in]{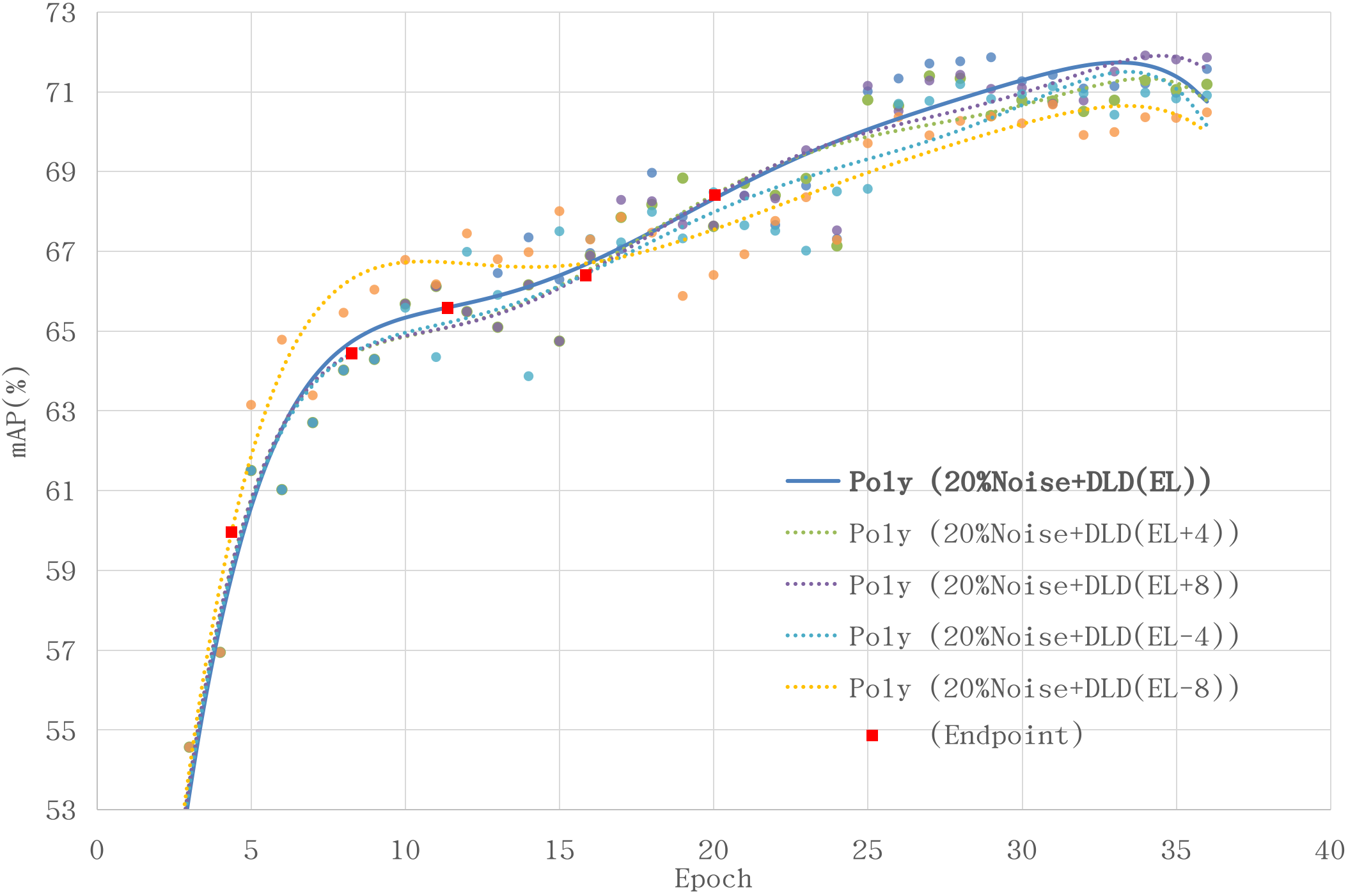}
            \label{val_map_slope}
            \caption{mAP\textbf{(Endpoints)} 20\% Noise}
	\end{minipage}
    \end{subfigure}
    \\ 
    \begin{subfigure}[b]{0.45\textwidth}
	\begin{minipage}[t]{\linewidth}
		\centering
		\includegraphics[width=3in]{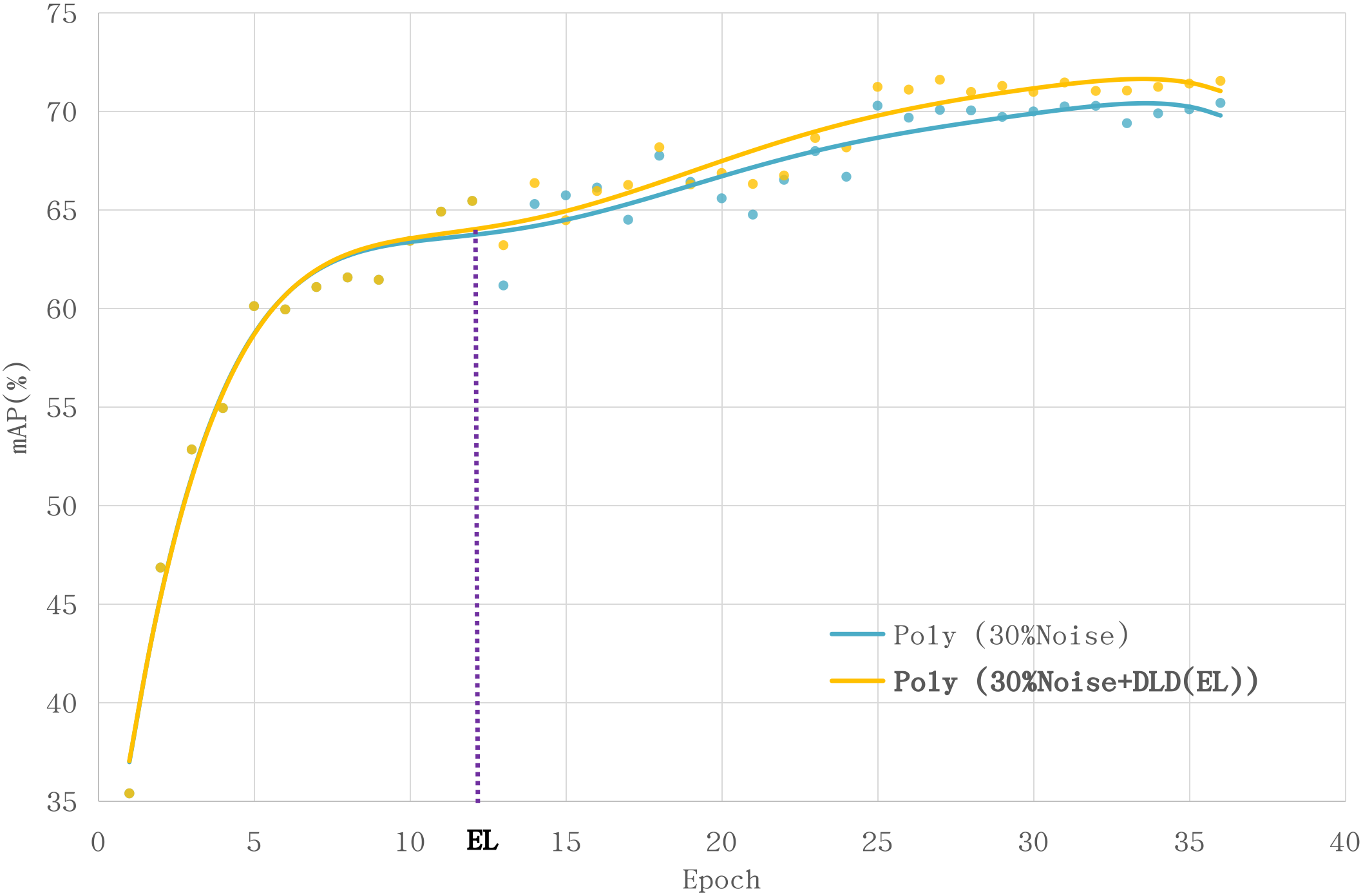}
            \label{fitted_acc}
            \caption{mAP\textbf{(Baseline vs. DLD)} 30\% Noise}
	\end{minipage}
    \end{subfigure}
    \begin{subfigure}[b]{0.45\textwidth}
	\begin{minipage}[t]{\linewidth}
		\centering
		\includegraphics[width=3in]{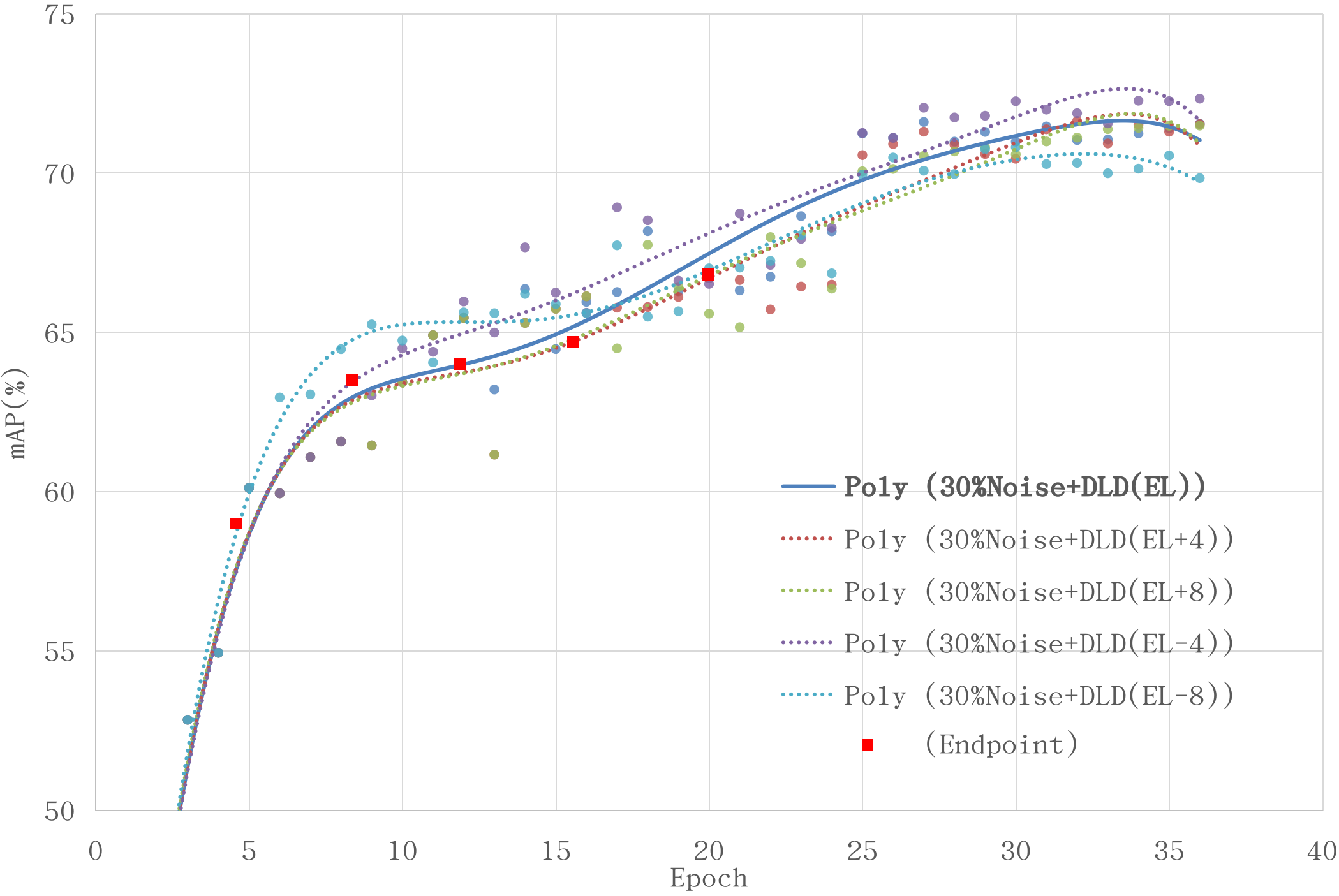}
            \label{acc_slope}
            \caption{mAP\textbf{(Endpoints)} 30\% Noise}
	\end{minipage}
    \end{subfigure}
    \\
    \begin{subfigure}[b]{0.45\textwidth}
	\begin{minipage}[t]{\linewidth}
		\centering
		\includegraphics[width=3in]{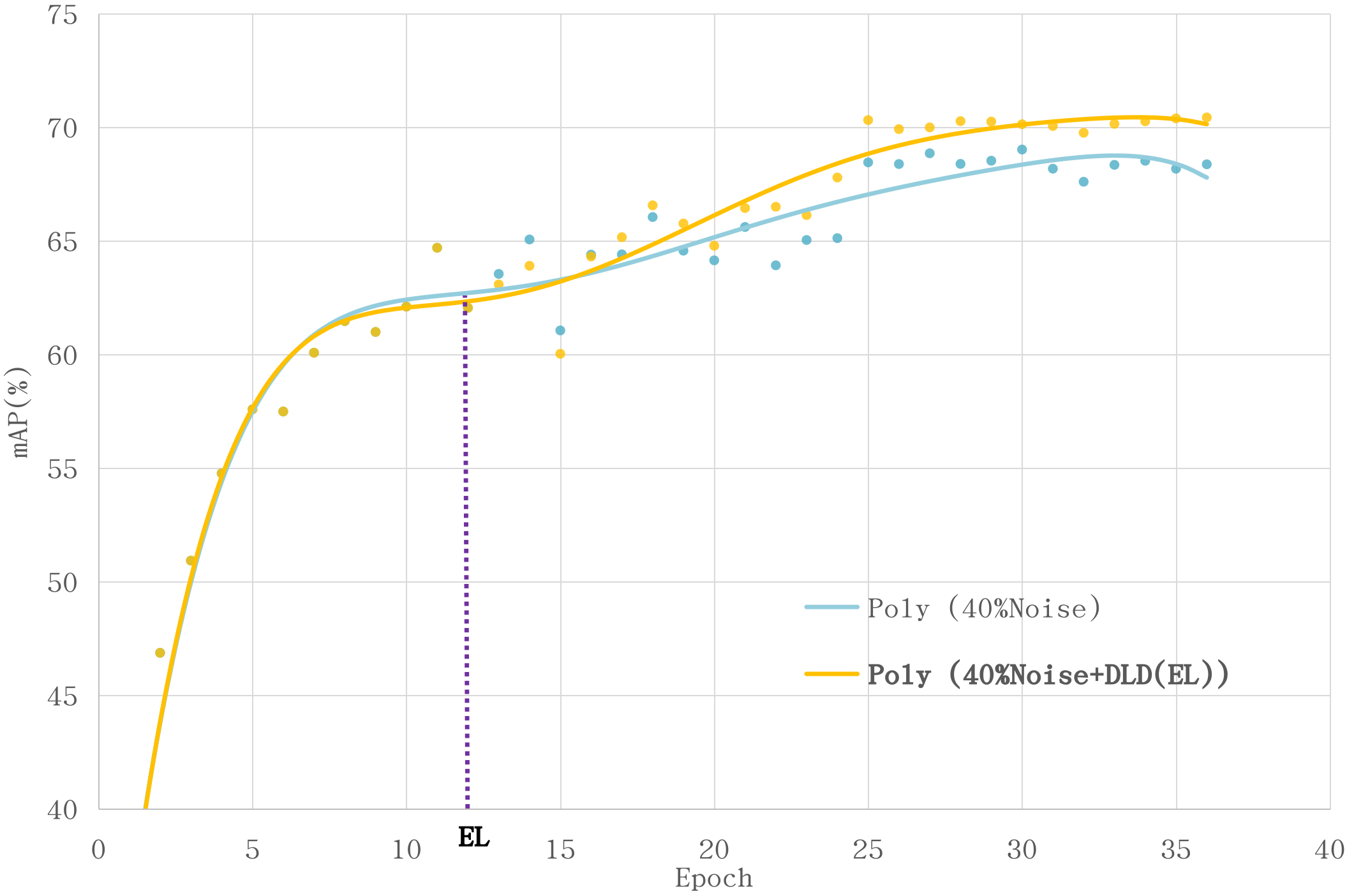}
            \label{fitted_acc}
            \caption{mAP\textbf{(Baseline vs. DLD)} 40\% Noise}
	\end{minipage}
    \end{subfigure}
    \begin{subfigure}[b]{0.45\textwidth}
	\begin{minipage}[t]{\linewidth}
		\centering
		\includegraphics[width=3in]{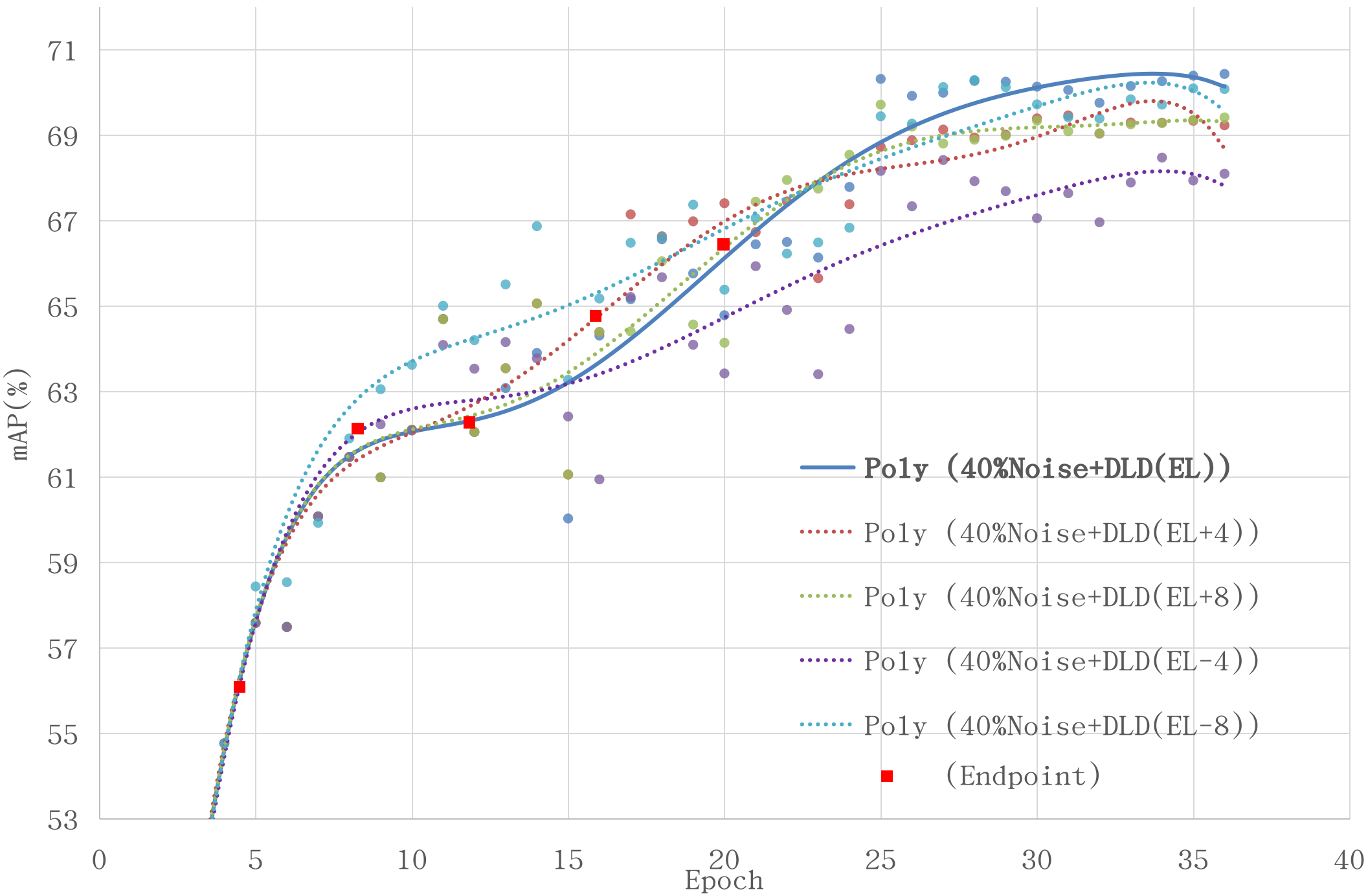}
            \label{acc_slope}
            \caption{mAP\textbf{(Endpoints)} 40\% Noise}
	\end{minipage}
    \end{subfigure}
    \caption{The dynamics of mean average precision (\textbf{mAP}) in the validation set of DOTA-v1.0 dataset, acquired by the  Oriented R-CNN with \textbf{LSKNet-Tiny(lsk-t)} backbone. The experiments are conducted on DOTA-v1.0 dataset contaminated with different level of category noises (20\%, 30\%, and 40\%). In the graphical, each line corresponds to a specific noise level scenario. The left segment of each line illustrates the fitted lines of \textbf{mAP values between the baseline and DLD} under the influence of noisy labels. Meanwhile, the right segment displays the \textbf{contrast of the fitted lines of mAP values for different endpoints}. The \textbf{endpoint} represents the specific point when DLD start to be applied.}
    \label{fig:val_el}
\end{figure*}

\begin{figure*}[h]
    \centering
    \begin{subfigure}[b]{0.9\textwidth}
	\begin{minipage}[t]{\linewidth}
		\includegraphics[width=6in]{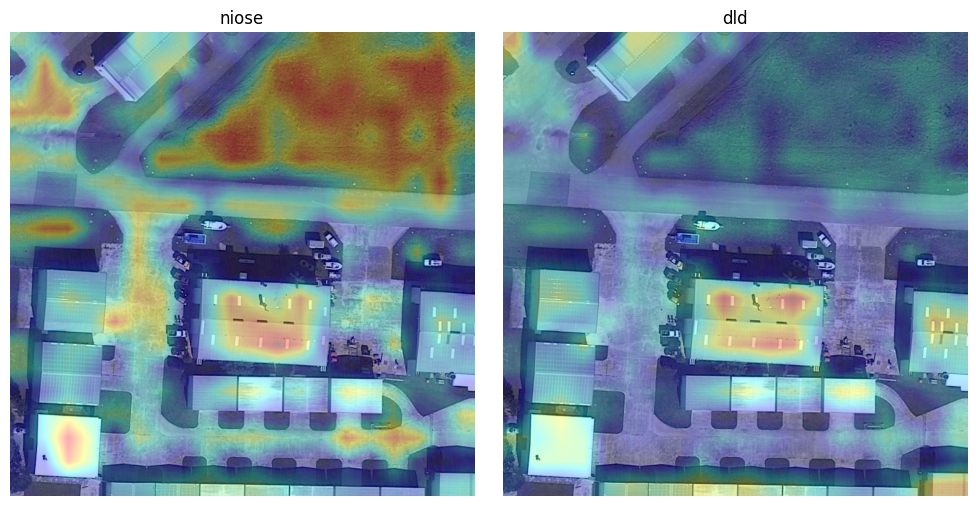} 
            \label{val_map}
            % \caption{mAP\textbf{(Baseline vs. DLD)} 20\% Noise}
	\end{minipage}
    \end{subfigure}
    \\ 
    \centering
    \begin{subfigure}[b]{0.9\textwidth}
	\begin{minipage}[t]{\linewidth}
		\includegraphics[width=6in]{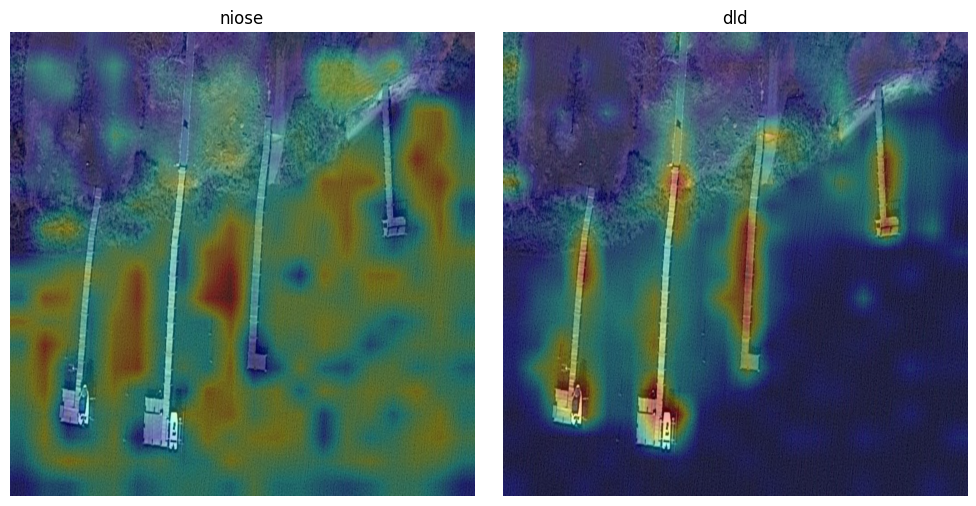} 
            \label{val_map}
            % \caption{mAP\textbf{(Baseline vs. DLD)} 20\% Noise}
	\end{minipage}
    \end{subfigure}
    \caption{Additional heatmap results on DOTA-v1.0 validation dataset, noise represents baseline trained with noise labels, dld represents baseline trained with noise labels and DLD.}
    \label{fig:heatmap1}
\end{figure*}

\begin{figure*}[h]
    \centering
    \begin{subfigure}[b]{0.9\textwidth}
	\begin{minipage}[t]{\linewidth}
		\includegraphics[width=6in]{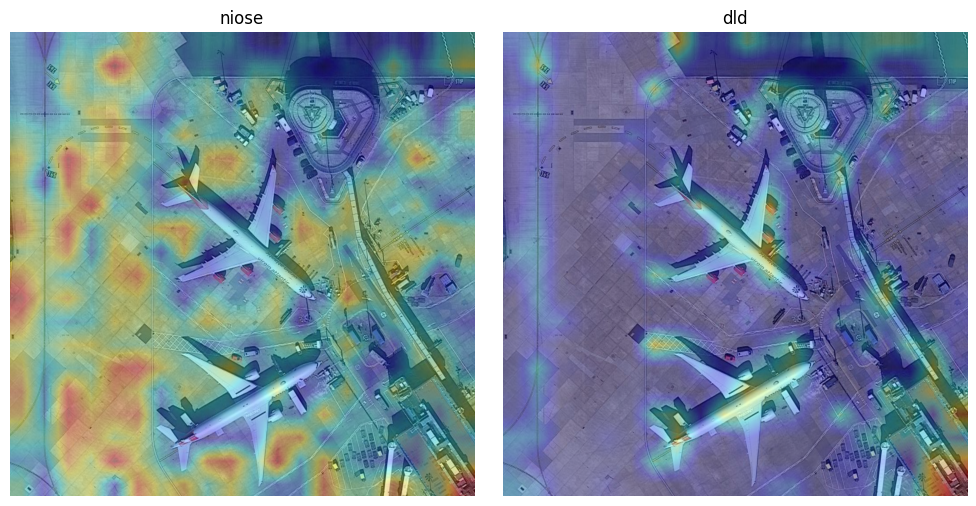} 
            \label{val_map}
            % \caption{mAP\textbf{(Baseline vs. DLD)} 20\% Noise}
	\end{minipage}
    \end{subfigure}
    \centering
    \begin{subfigure}[b]{0.9\textwidth}
	\begin{minipage}[t]{\linewidth}
		\includegraphics[width=6in]{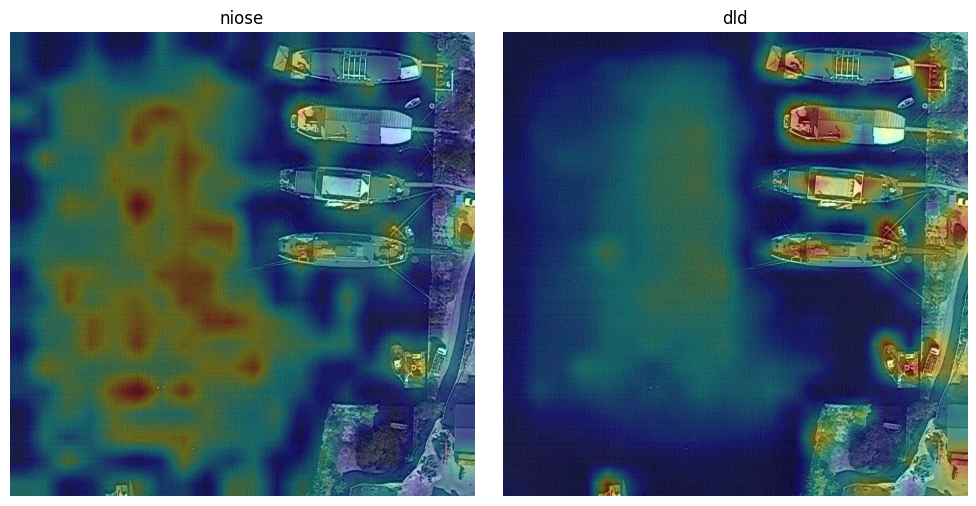} 
            \label{val_map}
            % \caption{mAP\textbf{(Baseline vs. DLD)} 20\% Noise}
	\end{minipage}
    \end{subfigure}
    \caption{Additional heatmap results on DOTA-v1.0 validation dataset, noise represents baseline trained with noise labels, dld represents baseline trained with noise labels and DLD.}
    \label{fig:heatmap2}
\end{figure*}

\begin{figure*}[ht]
\centering
\includegraphics[width=0.7\linewidth]{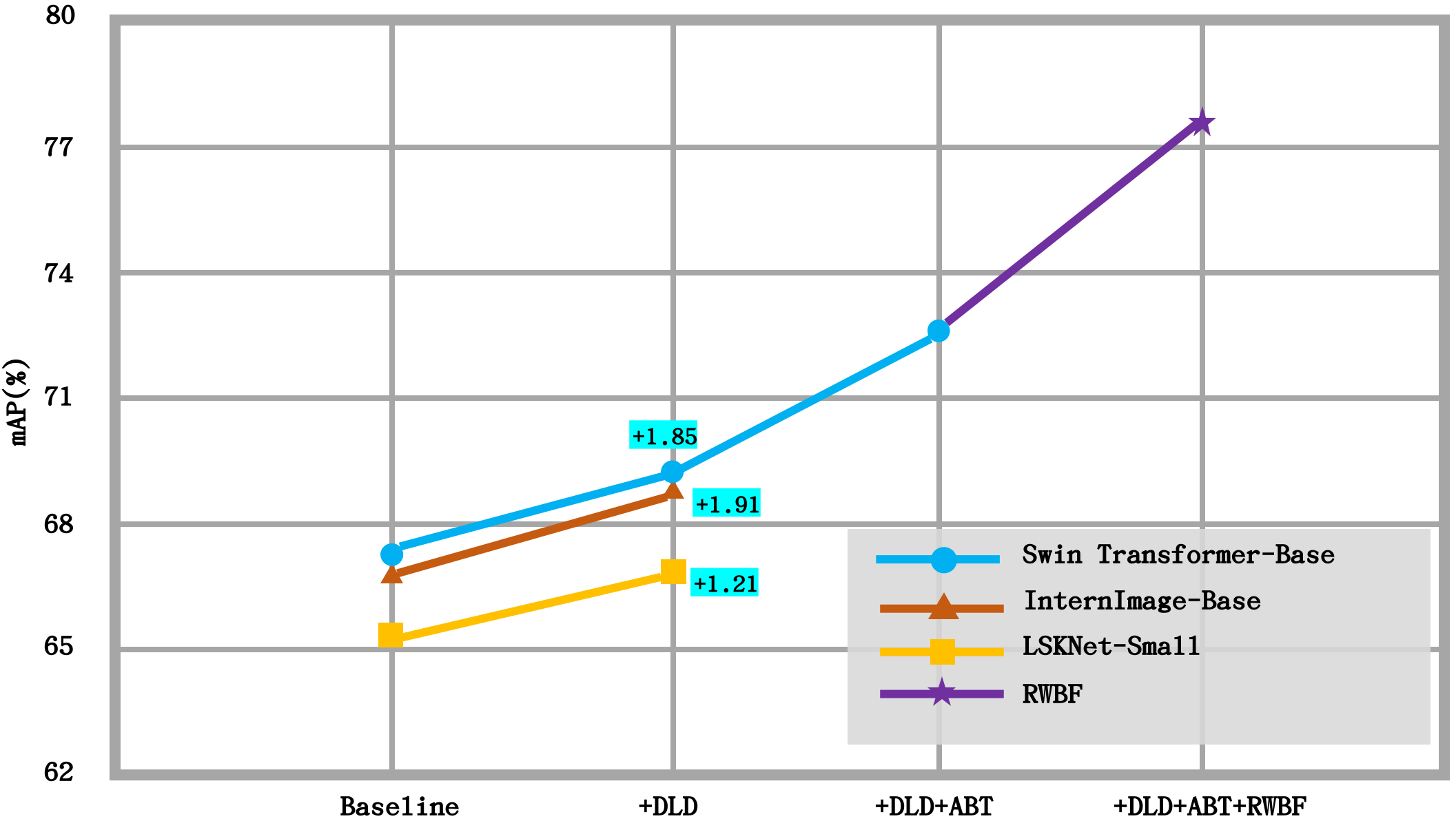}
\caption{Selected experiment results from the Final Round leaderboard of the 2023 National Big Data and Computing Intelligence Challenge.  Our approach(DLD), is presented combining with ABT (A Bag of Tricks) and RWBF (Rotation Weighted Bbox Fusion).}
\label{fig:competition}
\end{figure*}

\quad We include additional figures to complement the existing visuals. Figure \ref{fig:suppl_acc_map} provide supplementary information for Figure \ref{fig:fitted_map_acc_curve}, while Figure \ref{fig:heatmap1} and Figure \ref{fig:heatmap2} serves as a supplement to Figure \ref{fig:heatmap}.

% \onecolumn
% \begin{center}
\begin{onecolumn}
\begin{table}
\tablefirsthead{\hline Name & Demo & Number & Name  & Demo & Number  \\ \hline}
 % \tablelasttail{\hline}%最后一页最后一行的内容。
\begin{supertabular}{m{1.5cm}<{\centering}|m{3.5cm}<{\centering} | m{1.5cm}<{\centering} | m{1.5cm}<{\centering}|m{3.5cm}<{\centering} | m{1.5cm}<{\centering}} 
\hline
R1                        & \includegraphics[width=2cm]{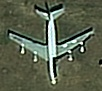} & 46   & L3  & \includegraphics{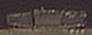}    & 104   \\ \hline
R2                        & \includegraphics{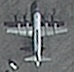} & 69   & S7  & \includegraphics{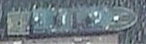}    & 153   \\ \hline
L2                        & \includegraphics{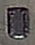} & 48   & T10 &  \includegraphics{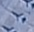}    & 107   \\ \hline
P6                        & \includegraphics[width=2.5cm]{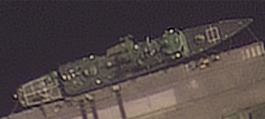} & 54   & M3  &  \includegraphics{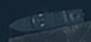}    & 107   \\ \hline
U2                        & \includegraphics[width=2cm]{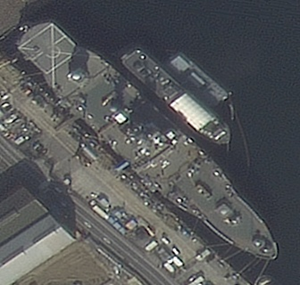} & 57   & A2  &  \includegraphics{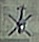}    & 108   \\ \hline
H3                        & \includegraphics{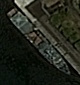} & 58   & I1  &  \includegraphics[width=2cm]{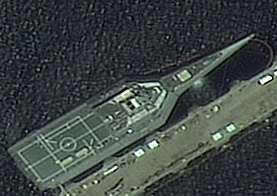}    & 140  \\ \hline
A6                        & \includegraphics[width=3cm]{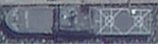} & 77   & C9  & \includegraphics[width=3cm]{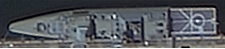}    & 223   \\ \hline
T11                       & \includegraphics{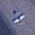} & 58   & C6  &  \includegraphics{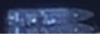}    & 120   \\ \hline
C5                        & \includegraphics[width=2cm]{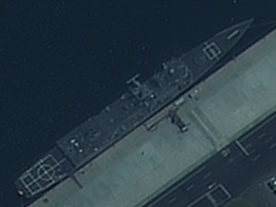} & 60   & A1  &  \includegraphics{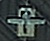}    & 129   \\ \hline
C7                        & \includegraphics[width=3cm]{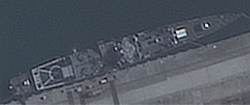} & 62   & C4  &  \includegraphics{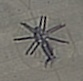}    & 131   \\ \hline
T1                        & \includegraphics{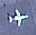} & 63   & U1  &  \includegraphics{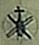}    & 132   \\ \hline
H2                        & \includegraphics[width=3cm]{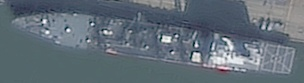} & 63   & T8  &  \includegraphics{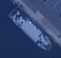}    & 133   \\ 
\end{supertabular}
\caption{This table presents 24 exemplar classes from the 2023 National Big Data and Computing Intelligence Challenge dataset. It encompasses a comprehensive array of 98 ship and airplane categories. The "Name" column represents the category, "Demo" displays image examples for each category, and "Number" indicates the instance number within each category.}
\label{traning data of competiton}
\end{table}
\end{onecolumn}

\end{document}